\documentclass{article}

\usepackage{arxiv}

\usepackage{graphics}
\usepackage{url}
\usepackage{graphics}
\usepackage[english]{babel}
\usepackage{hyperref}
\usepackage{listings}
\usepackage{color}
\usepackage{enumitem}
\usepackage{mathtools}
\usepackage{amsmath}
\usepackage{xcolor}
\usepackage{graphicx}
\usepackage{booktabs}
\usepackage{float}
\usepackage{cleveref}
\usepackage{amsthm}
\usepackage{svg}
\usepackage{siunitx}
\usepackage{tabularx} 
\usepackage{amssymb}

\usepackage[utf8]{inputenc} % allow utf-8 input
\usepackage[T1]{fontenc}    % use 8-bit T1 fonts
\usepackage{booktabs}       % professional-quality tables
\usepackage{amsfonts}       % blackboard math symbols
\usepackage{nicefrac}       % compact symbols for 1/2, etc.
\usepackage{microtype}      % microtypography
\usepackage{lipsum}
\usepackage{graphicx}
\usepackage{caption}
\usepackage{ctable}
\usepackage{csquotes}
\usepackage[numbers]{natbib}
\graphicspath{ {./images/} }

\usepackage{titlesec}

\titleformat{\paragraph}
{\normalfont\normalsize\bfseries}{\theparagraph}{1em}{}
\titlespacing*{\paragraph}
{0pt}{3.25ex plus 1ex minus .2ex}{1.5ex plus .2ex}

\newcommand{\norm}[1]{\left\lVert#1\right\rVert}

\newcommand{\citea}[1]{\citeauthor{#1} \cite{#1}}
\newcommand{\citepa}[1]{(\citeauthor{#1} \cite{#1})}

\newtheorem{Definition}{Definition}
\numberwithin{Definition}{section}

\usepackage{array}
\newcolumntype{L}[1]{>{\raggedright\let\newline\\\arraybackslash\hspace{0pt}}m{#1}}
\newcolumntype{C}[1]{>{\centering\let\newline\\\arraybackslash\hspace{0pt}}m{#1}}
\newcolumntype{R}[1]{>{\raggedleft\let\newline\\\arraybackslash\hspace{0pt}}m{#1}}

\title{Anomaly Detection in Univariate Time-series: A Survey on the State-of-the-Art}

\author{Mohammad Braei\\Department of Computer Science\\Technische Universität Darmstadt\\mohammad.braei@stud.tu-darmstadt.de \And  Dr.-Ing. Sebastian Wagner \\Telecooperation Group\\Technische Universität Darmstadt\\s.wagner@tk.tu-darmstadt.de }
\begin{document}
\maketitle
\begin{abstract}
Anomaly detection for time-series data has been an important research field for a long time. Seminal work on anomaly detection methods has been focussing on statistical approaches. In recent years an increasing number of machine learning algorithms have been developed to detect anomalies on time-series. Subsequently, researchers tried to improve these techniques using (deep) neural networks. In the light of the increasing number of anomaly detection methods, the body of research lacks a broad comparative evaluation of statistical, machine learning and deep learning methods. This paper studies 20 univariate anomaly detection methods from the all three categories. The evaluation is conducted on publicly available datasets, which serve as benchmarks for time-series anomaly detection. By analyzing the accuracy of each method as well as the computation time of the algorithms, we provide a thorough insight about the performance of these anomaly detection approaches, alongside some general notion of which method is suited for a certain type of data. 
\end{abstract}

% keywords can be removed

\section{Introduction}
Detecting anomalies has been a research topic for a long time. In a  world of digitization, the amount of data transferred exceeds the human ability to study it manually. Hence, automated data analysis becomes a necessity. One of the most important data analysis tasks is the detection of anomalies in data. Anomalies are data points which deviate from the normal distribution of the whole dataset, and anomaly detection is the technique to find them. \\
The impact of an anomaly is domain-dependent. In a dataset of network activities, an anomaly can imply an intrusion attack. An anomaly in a financial transaction can hint on financial fraud,  anomalies in medical images can be caused by diseases. Other objectives of anomaly detection are industrial damage detection, data leak prevention, identifying security vulnerabilities or military surveillance. \\
Anomaly detection methods are specific to the type data. For instance, the algorithms used to detect anomalies in images are different to the approaches used on data streams. This paper focuses on methods for anomaly detection in time-series data. Anomaly detection on time-series has been a long-time topic of interest. In 1979, \citeauthor{ExploratoryDataAnalysis} \cite{ExploratoryDataAnalysis} proposed a statistical approach to detect anomalies on time-series. \citeauthor{EstimationTimeSeriesParametersOutliers} \cite{EstimationTimeSeriesParametersOutliers} proposed to use the likelihood ratio test (LRT) to detect anomalies on time-series.\\
With the ever-growing computational power in recent decades, machine learning approaches increased in popularity for data science tasks such as classification and pattern detection. Therefore, many researchers started to use these machine learning methods to detect anomalies in time-series. For instance, clustering methods such as k-Means can be used to detect anomalous points in time-series data.

In the last decade, deep learning approaches achieved tremendous progress in computer vision tasks. This success motivated researchers to leverage these methods to detect anomalies. Various deep learning approaches such as Multi -Layer Perceptrons (MLPs), Convolution Neural Network (CNNs) and Long-Short Term Memory (LSTMs) were proposed as anomaly detection techniques. 
While there is a wide spectrum of anomaly detection approaches today, it becomes more and more difficult to keep track of all the techniques. As a matter of fact, it is not clear which of the three categories of detection methods, i.e., statistical approaches, machine learning approaches or deep learning approaches is more appropriate to detect anomalies on time-series data. To the best of our knowledge, there is no study comparing approaches of these three categories in their accuracy and performance.

This paper presents a quantitative comparison of multiple approaches of each category. We select a wide range of methods to cover well-performing techniques of each class. One of the main contributions of this work is that it focuses on time-series data. In order to provide a reliable comparison, the methods are evaluated on multiple time-series datasets. \\
This paper is structured as follows:
\begin{enumerate}
    \item \textbf{Basics:} In this section, the main concepts of anomaly detection on time-series, which are fundamental to the subject, are defined. These concepts are vital to understand the algorithms in the following sections.
    \item \textbf{Selected Anomaly detection approaches for time series:} This section introduces different anomaly detection algorithms of the three main categories. 
    \item \textbf{Approach:} While first introducing related and similar works, here we explain how the evaluation of the different methods is carried out. 
    \item \textbf{Experiments:} As a preceding step to the evaluations, this section lists the setup of the experiments by listing all hyperparameters of the used algorithms.
    \item \textbf{Results:} This section illustrates and discusses the evaluation results.
    \item \textbf{Conclusion:} The last section concludes the paper while also providing an insight into future work.
\end{enumerate}{}

\section{Foundations}
\label{chapter:basics}
In this section we define basic concepts which are fundamental in the anomaly detection process. 
\subsection{Anomalies and outliers}
There is no consent about the distinction of anomalies and outliers. 
On one hand, the following citation is mostly referenced to prove the equality of anomaly and outliers:
\begin{quote}
“\textit{Outliers are also referred to as abnormalities, discordants, deviants, or anomalies in the data mining and statistics literature.}” -- \citeauthor{OutlierAnalysisAggarwal2016OA3086742} \cite{OutlierAnalysisAggarwal2016OA3086742}
\end{quote}
On the other hand, there are definitions which regard outliers as a broader concept, which also includes noise in addition to anomalies \cite{Salgado2016}. Others consider outliers as corruption in data, while anomalies are irregular points, but with a specific pattern \cite{RobustMultivariateAutoregressionforAD}.\\
As we are evaluating time-series in this paper, we consider the two terms, outlier and anomaly, interchangeably. 

The important point is to deliver a formal definition for the concept of anomaly. This is essential, because different definitions of anomalies imply varying methods to detect them. Thus, it is necessary to define the main characteristics of anomalies and highlight the boundaries by the definition.
The most common definition of anomalies is the following:
\begin{quote}
    “\textit{Anomalies are patterns in data that do not conform to a well defined notion of normal behavior.}” -- \citea{Chandola2009}
\end{quote}
Chronologically, one of the first definitions was given by \citea{doi:10.1080/00401706.1969.10490657}. He defined outliers in 1969 as: 
\begin{quote}
    “\textit{An outlying observation, or "outlier," is one that appears to deviate markedly from other members of the sample in which it occurs.}”
\end{quote}
\citea{RePEc:eee:intfor:v:12:y:1996:i:1:p:175-176} used the following definition: 
\begin{quote}
    “\textit{An  observation  (or  subset of  observations)  which  appears to  be inconsistent with the remainder of that set of data.}”
\end{quote}
And finally \citea{GVK02435757X} defines outliers as follows: 
\begin{quote}
    “\textit{An outlier is an observation which deviates so much from the other observations as to arouse suspicions that it was generated by a different mechanism.}”
\end{quote}{}
All these definitions highlight two main characteristics of anomalies:
\begin{itemize}
    \item The distribution of the anomalies deviates remarkably from the general distribution of the data.
    \item The big majority of the dataset consists of normal data points. The anomalies form only a very small part of the dataset.
\end{itemize}
These two aspects are fundamental to the development of anomaly detection methods. Especially the second property prevents us from using common classification approaches that rely on balanced datasets, and enables us to use approaches like auto-encoders as a semi-supervised method to detect anomalies, which will be explained in the following sections. \\
Thus, in this paper anomaly and outliers are defined as follows:
\begin{Definition}\label{def:DefinitionAnomaly}
An anomaly is an observation or a sequence of observations which deviates remarkably from the general distribution of data. The set of the anomalies form a very small part of the dataset. 
\end{Definition}{}

It is also important to distinguish between anomalies and noise. Noise can be a mislabeled example (class noise) or errors in the attributes of the data (attribute noise) \cite{Salgado2016} which are not of interest to data analysts \cite{OutlierAnalysisAggarwal2016OA3086742}. While for instance, in a set of medical images an anomaly may show a tumor, noise is just a random variation of brightness and color information which is unwanted. Thus, noise is not of interest to the analyst, while an anomaly is.\\
Whereas the difference between anomalies and noise is highlighted here, it still remains a difficult task to differentiate between them in some sort of data. This is illustrated in Figures \ref{fig:WithoutNoise} and \ref{fig:WithNoise}:

\begin{figure}[H]
\centering
  \captionsetup{justification=centering,margin=0cm}
\begin{minipage}{.5\textwidth}
        \centering
 \includegraphics[scale=0.5]{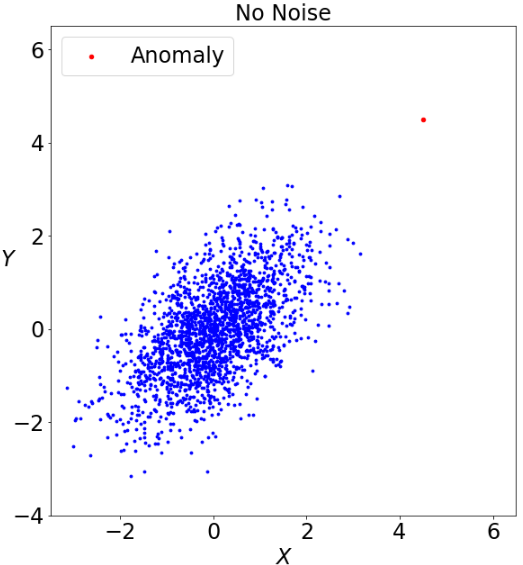}
 \caption{Data without noise.}
 \label{fig:WithoutNoise}
\end{minipage}%
\begin{minipage}{.5\textwidth}
        \centering
 \includegraphics[scale=0.5]{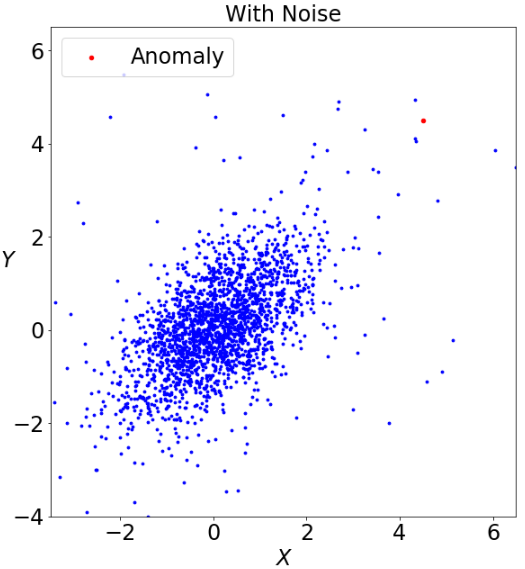}
 \caption{Data including noise.}
 \label{fig:WithNoise}
\end{minipage}
\end{figure}

In both figures the data distribution is the same. In Figure \ref{fig:WithoutNoise} the anomalous point marked in red seems to be obvious as it deviates significantly from the rest. But in Figure \ref{fig:WithNoise} it is difficult to distinguish the anomaly point from the other points in the sparse space. 
This example shows that the difficulty to distinguish  between anomalies and noise depends on the dataset. Thus, a deep understanding of the dataset is necessary to do so.

Anomalies should also be distinguished from novelties \cite{Chandola2009,PIMENTEL2014215}. Novelty patterns are data points which have not been observed in the data yet. The difference to anomalies is, that novelties are considered \textit{normal} after being detected once. For example, a new communication pattern to a server after implementing a new protocol. However, due to the fact that most methods used for novelty detection are also used for anomaly detection and vice versa, in this paper we treat them equally.

\subsection{Types of Anomalies}\label{AnomalyTypes}
Anomalies can appear in different forms. Commonly, three different types of anomalies exist:
\begin{enumerate}
\item \textbf{Point anomalies:} If a point deviates significantly from the rest of the data, it is considered a point anomaly. For example, a big credit transaction which differs from other transactions is a point anomaly. Hence, a point $X_t$ is considered a point anomaly, if its value differs significantly from all the points in the interval $[X_{t-k}, X_{t+k}],k\in\mathbb{R}$ and $k$ is sufficient large.%\todo[inline]{Was ist k?-A: Habe es ueberarbeitet}.
\item \textbf{Collective anomalies:} There are cases where individual points are not anomalous, but a sequence of points are labeled as an anomaly. For example, a bank customer withdraws \$500 from her bank account every day of a week. Although withdrawing \$500 occasionally is normal for the customer, a sequence of withdrawals is an anomalous behavior.
\item \textbf{Contextual anomalies:} Some points can be normal in a certain context, while detected as anomaly in another context: Having a daily temperature of \ang{35} C in summer in Germany is normal, while the same temperature in winter is regarded as an anomaly.
\end{enumerate}
Knowing a-priori, which kind of anomaly the data might contain, assists the data analyst to select the appropriate detection method. Some approaches that are able to detect point anomalies fail to identify collective or contextual anomalies altogether.

\subsection{Stochastic Processes and Time-series}
\label{SectionTimeSeries}
\begin{figure}[H]
 \centering
  \captionsetup{justification=centering,margin=2cm}
 \includegraphics[scale=0.5]{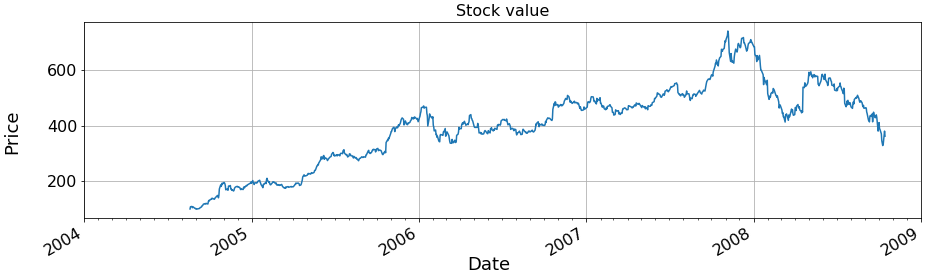}
 \caption{Sample time-series showing the prices of a sample stock over five years}
 \label{fig:SampleTimeSeries}
\end{figure}

The data that is analyzed in this paper are time-series. Thus, it is fundamental to provide a definition of it. Primarily, another term has to be defined, which regularly occurs simultaneously with time series: \textit{Stochastic Process}. \citeauthor{TimeSeriesAnalysisUnivariateMultivariateMethods} \cite{TimeSeriesAnalysisUnivariateMultivariateMethods} provides a comprehensive definition:
\begin{Definition}\label{def:StochasticProces}
A stochastic process is a family of time indexed random variables $Z(\omega,t)$, where $\omega$ belongs to a sample space and $t$ belongs to an index set.
\end{Definition}{}
Hence, if $t$ is fixed then $Z$ is a random variable over the sample space. \\
A \textit{time series} is a realization of a certain stochastic process. A formal distinct definition for time-series is as follows:
\begin{Definition}
\label{def:DefinitionTimeSeries}
A time-series is a sequence of observations taken by continuous measurements over time. Generally, the observations are picked up in equidistant time intervals:
$T=(t_0^d,t_1^d,...,t_t^d), d\in \mathbb{N}_+, t\in \mathbb{N}\text{ where }d \text{ defines the dimension of time series}$
%$d$ defines the dimension of time-series.
\end{Definition}{}
 A time-series can be a sequence of observations from one source, i.e., one sensor. In this case $d=1$ and the series is univariate. If we collect information from more than one sensor, $d>1$,  we have a multivariate time-series. In this paper, we consider only \textbf{discrete} time-series, therefore $t \in \mathbb{N}$, perceived in equal time intervals.\\ 
Instances of univariate time-series are cash transactions or weather histories. Figure \ref{fig:SampleTimeSeries} shows a time plot of a sample time-series of an observed stock market value of an asset over five years.

An example of a multivariate time-series is the collected data from several sensors installed in a car. \\
One main difference between time-series and other datasets is that the observations  do not only depend on components $d$, but also on the time feature $n$. Thus, time-series analysis and the used statistical methods are mostly different from the methods used for random variables that assume independence and constant variance of the random variables.\\
To data analysts, time-series are important in a variety of fields like economy, healthcare and medical research, trading, engineering and geophysics. These data are used for forecasting and anomaly detection. 

\subsection{Time-series patterns}
Time-series have some important properties which we will define briefly. They are significant to the statistical anomaly detection methods used later.
\subsubsection{Trend}
A time-series has a trend, if its mean $\mu$ is not constant, but increases or decreases over time. A trend can be linear or non-linear. The time-series in Figure \ref{fig:SampleTimeSeries} has a positive trend from 2005 until 2008, and a negative trend afterwards.

\subsubsection{Seasonality}
\begin{figure}[H]
     \centering
      \captionsetup{justification=centering,margin=2cm}
     \includegraphics[scale=0.5]{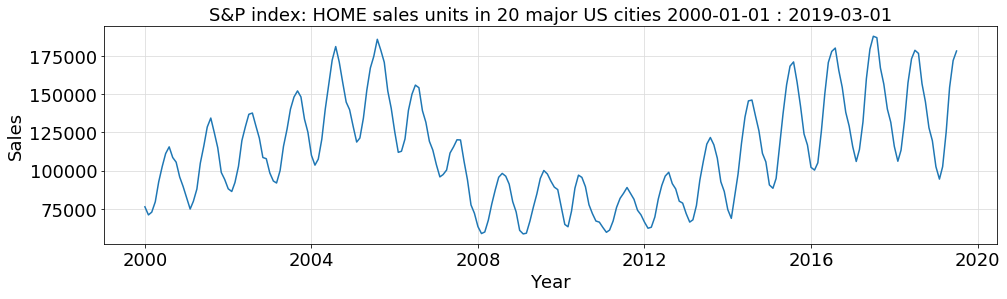}
     \caption{Sample time-series showing the prices of a stock over five years}
     \label{fig:SeasonalData}
    \end{figure}

Seasonality is the periodic recurrence of fluctuations. The time-series is called seasonal because seasonal factors like time of the year or day of the week, or other similarities are influencing it. Thus, it always has a fixed period of time that is limited to a year. Figure \ref{fig:SeasonalData} shows a seasonal time-series. It is the monthly home sales index for 20 major US cities between the years 2000 and 2019 \cite{SnPDowJonesIndicesLLC}.    

\subsubsection{Cycles}
A cyclic time-series is influenced by time factors where the period is not fixed and the duration is above a year, e.g., a decade. The time-series in Figure \ref{fig:SeasonalData} also has an approximate 12 year cycle.

\subsubsection{Level}
The time-series level is equal to the mean of the series. If a time-series has a trend, then it is often said that the level is changing.

\subsubsection{Stationarity}
Intuitively, a stationary time-series is a time-series having the same characteristics over every time interval. Formally, we can express it as follows \cite{HyndmanforecastingPrinciplesAndPractice}: 
\begin{Definition}
${X_t}$ is a stationary time-series, if  $\text{ } \forall s \in \mathbb{R}$: the distribution of $(x_t,...,x_{t+s})$ is equal.
\end{Definition}{}
The above definition implies that a stationary time-series ${x_1,..,x_T}$ will have the following characteristics:
    
    \begin{enumerate}
        \item \textbf{Constant mean}, thus no trend exists in the time-series.
        \item The time-series has a \textbf{constant variance}.
        \item There is a \textbf{constant autocorrelation} over time.
        \item The time-series has \textbf{no seasonality}, i.e., no periodic fluctuations. 
    \end{enumerate}{}

\subsubsection{White noise}
White noise $\epsilon_t$ is a stochastic process, which is uncorrelated over time from a fixed distribution with a constant mean $\mu=0$ and a constant and finite variance $\sigma$. Thus, white noise is stationary. One important characteristic of white noise is that its autocorrelation function (ACF) and its partial autocorrelation function (PACF) are zero, meaning that there is no dependence between two timestamps. Usually, in many theoretical models it is assumed that white noise is Gaussian: $\epsilon\sim \mathcal{N}(0,\,\sigma)\,$.

\subsection{Anomaly detection}\label{AnomalyDetectionMethod}
After having a general definition for anomaly and time-series, we will define what anomaly detection means and what kind of methods exist.\\
In literature, different terms are used that have the same or similar meaning to \textit{Anomaly Detection}: \textit{Event detection, novelty detection, (rare) event detection, deviant discovery, change point Detection, fault detection,  intrusion detection or misuse detection} \cite{OutlierDetectionForTemporalData}. The different terms reflect the same objective: to detect rare data points that deviate remarkably from the general distribution of the dataset. The amount of deviation is usually regarded as a measure of strength of the anomaly or---probabilistically--- the likelihood of being an anomaly, which is called \textit{anomaly score}. Thus formally, anomaly detection can be defined as a function $\phi$:

    \begin{equation}\label{EquationAnomalyDetection}
        \begin{split}
             \phi : \mathbb{R}^n & \to \mathbb{R}\\
             \phi(x) & \mapsto \gamma 
     \end{split}{}
    \end{equation}{}
where $\gamma$ is the anomaly score, $x \in X \subseteq \mathbb{R}^n$, and $X$ is the dataset. 
\\
To convert the continuous value $\gamma$ into a binary label \--- normal vs. anomaly \--- a threshold $\delta \in \mathbb{R}$ is defined where all points with an anomaly score greater than $\delta$ are marked as an anomaly. Thus, let $\phi_{score}:= \phi$, then the binary labeling anomaly detection method $\phi_{binary}$ can be defined as:
    \begin{equation}\label{EquationAnomalyDetectionBinary}
        \begin{array}{l}
             \phi_{binary} : \mathbb{R}^n \to \{normal,anomaly\}\\\\
            
            \phi_{binary}(x) \mapsto
            \begin{cases*}
            anomaly, & if $\phi_{score}(x)>\delta$\\
            normal,       & otherwise
            \end{cases*}
        \end{array}{}
    \end{equation}{}

Anomaly detection using $\phi_{binary}$ is not a trivial binary classification. As stated in Definition \ref{def:DefinitionAnomaly}, anomalies form a very small part of the dataset. Often the anomalous part of a dataset is less then 1\%. Therefore, usual binary classifiers would achieve above 99\% accuracy if all data points would be labeled as normal, making anomaly detection a more difficult task. Nevertheless, to achieve satisfying results in anomaly detection, the proper anomaly detection method has to be selected which is dependent on the properties of the inspected data. The following properties are important for selecting the appropriate approach:
\begin{enumerate}
    \item Temporal vs Non-Temporal data: Non-Temporal data can be medical images, protein sequences etc. Temporal data include time-series, but also data with timestamps of unequal interval. 
    \item Univariate vs Multivariate data: Univariate data takes only one dimension, e.g. the stock price, while multivariate data contains multiple dimensions. Instances of multivariate data are images or time-series observed by several sensors.
    \item Labeled or unlabeled data: A dataset is labeled if an annotation exists for each element in the dataset, which determines if it is a normal or anomalous data point. A labeled dataset with normal and anomalous points is the object of supervised anomaly detection methods. It is possible that the dataset is completely labeled but only consists of normal points. Then, it can be analysed by semi-supervised methods. Finally, unlabeled data is the object of unsupervised anomaly detection methods.
    \item Types of anomalies in the dataset: Section \ref{AnomalyTypes} introduced different anomaly types. This information affects the selection of the anomaly method. Point anomalies are detected by methods for rare classification. To detect collective anomalies, the methods focus on unusual shapes in the data, while searching for deviation. This aids in finding contextual anomalies.
    
\end{enumerate}{}

In this paper, we focus on univariate temporal data and specifically time-series containing labeled normal and anomalous points. Therefore, we will introduce the general concepts of these kinds of methods here.

\subsubsection{Anomaly detection on time series}
Anomaly detection on non-temporal data like spatial data is different than on time-series. For example one of the main methods to detect anomalies in spatial data is by measuring the deviation of the abnormal points to the rest of the data. Another way is to cluster the whole dataset and mark all points as anomalies that lie in less dense regions. The main assumption about spatial data is that the data points are independent from each other. 

This is different in time-series data. Here, the data points are not completely independent, but it is assumed that the latest data points in the sequence influence their following timestamps. Following this, values of the sequence change smoothly or show a regular pattern. Thus, sudden changes in the sequence will be regarded as an anomaly. To show this behavior, consider the following example which demonstrates a time-series listing the temperature of an engine recorded every 10 minutes:
\ang{30}, \ang{31}, \ang{33}, \ang{32}, \ang{34}, \ang{35}, \ang{35}, \ang{85}, \ang{87}, \ang{88}, \ang{89}, \ang{89}.
If these points are regarded as independent points, most methods will not identify any anomalous behavior, but detect two equally distributed clusters. But in a time-series, the sudden change from \ang{35}C to \ang{85}C should be detected as an anomaly. The dependency between timestamps also results in the fact that anomalies in time-series are generally contextual or collective. \\
\citea{OutlierAnalysisAggarwal2016OA3086742} breaks down anomaly detection methods for time-series into two main categories:
\begin{enumerate}
    \item Anomaly detection based on prediction of the time series
    \item Anomaly detection based on unusual shapes of the time series
\end{enumerate}{}
Most statistical anomaly detection methods on time-series are based on time-series prediction. On the other side, there are several machine learning methods, which try to detect anomalies using clustering methods on time-series. The selected method is dependent on whether the time-series is univariate or multivariate. As the focus of this paper are univariate time-series, we will provide an overview of their characteristics.

\subsubsection{Anomaly detection on univariate time series}
Anomaly detection in time-series is strongly linked to time-series analysis and forecasting methods. To detect anomalies in univariate time-series, a forecasting model is fitted to the training data. Then, the test data is used to make predictions. To make a prediction on the test data, usually a sliding window is used. A sliding window is a subsequence of a time-series, which is fed as the input to the model enabling it to predict the value for following timestamp. Formally, let $w$ be the width of the sliding window, and suppose we want to predict $x_i$, and $\psi$ is the forecasting model. To forecast $x_i$ the following function and input data is used:
    \begin{equation}
        \begin{array}{l}
             \psi : \mathbb{R}^w \to \mathbb{R}\\
            \hat{x}_i = \psi((x_{i-w},...,x_{i-1}))
            
        \end{array}{}
    \end{equation}
% \begin{equation}
%     \hat{x}_i = \psi((x_{i-w},...,x_{i-1}))
% \end{equation}{}
The anomaly score can be computed by measuring the distance between the predicted value $\hat{x}_i$ and the real value $x_i$:
\begin{equation}
    e_i = d(x_i,\hat{x}_i)
\end{equation}{}
where $d$ is a distance function. In univariate time-series, usually the euclidean distance is used. The deviation $e_i$ \--- also called error value \--- is proportional to the anomaly score. If the anomaly score is above a threshold $\delta \in \mathbb{R}$, it is marked as an anomaly.  \\
As mentioned, there are also approaches which try to detect anomalies in time-series by looking for unusual shapes by using machine learning approaches. For instance, \citea{OneClassSVMAnomalyDetectionCommunicationNetworkperformanceData} use One-Class Support Vector Machines to detect anomalies in time-series.%\todo[inline]{beispiel als referenz?-A:Ueberarbeitet} 
Therefore, in contrast to spatial data, a sliding window with width $w$ is defined. Then, for each timestamp $x_i$ the preceding $w$-timestamps are analysed using some clustering or density methods. These methods are based on the assumption, that an contextual or collective anomaly will show a deviating shape which can be detected by these clustering or density methods. We will return to this issue in Section \ref{chapter:chapterAnomalyDetectionApproache}.

\subsubsection{Supervised vs. Semi-supervised vs. Unsupervised Anomaly detection methods}
If the time-series dataset is labeled, such that for each timestamp it is known if it is an anomaly or not, and additionally the dataset contains normal and anomalous timestamps, then a supervised anomaly detection method can be used. Supervised anomaly detection methods are able to detect an appropriate value for $\delta \in \mathbb{R}$ to classify all timestamps $x_i$ as an anomaly if the corresponding anomaly score is $\phi(\hat{x_i})> \delta$.\\
Semi-supervised approaches can be used if the dataset only consists of normal points and no anomaly exists. Then a model is trained, which fits to the distribution of the time-series and detects any new point deviating from this distribution as an anomaly. One-Class SVN, autoencoders or GANs are usual methods used for this sort of data.\\
Finally, unsupervised anomaly detection methods assume that the time-series data is unlabeled. Most unsupervised anomaly detection methods try to determine $\delta$ by analyzing the distribution of all $e_i, i\in\{1,...,N\}$ and use the  $\tau$-percentile value as $\delta \in \mathbb{R}$. One widespread approach is to set $\delta = 3\sigma$ where $\sigma$ is the standard deviation of the distribution of $e_i, i\in\{1,...,N\}$.\\
In this paper we will focus on supervised anomaly detection methods and use labeled datasets in our experiments to evaluate them. 

\subsubsection{Statistical vs. Machine Learning vs. Deep Learning Anomaly detection approaches on time series}
\citea{DeepANT_DeepLearningUnsupervisedAnomalyDetectionTimeSeries} categorize outlier detection methods in probabilistic models, statistical models, linear models, proximity based models, and outlier detection in high dimensions while referencing to Aggarwal's book \cite{OutlierAnalysisAggarwal2016OA3086742}.\\
We believe that all anomaly detection methods on time-series data can be divided in three main categories:
\begin{enumerate}
    \item Statistical methods
    \item Classical machine learning methods
    \item Methods using neural networks (Deep Learning)
\end{enumerate}{}
This categorization is goal-driven as we want to inspect if they behave differently. Some studies merge the second and third class in machine learning approaches \cite{StatisticalMachineLeanringForecastingMethodsConsernsWaysForward}. The boundary of the third category using neural networks is rather clear as it only contains methods using some kind of a neural network. In Section \ref{chapter:chapterAnomalyDetectionApproache} we will define what a neural network is. In contrast, the boundary between statistical and machine learning approaches are vague. Generally, statistical approaches assume that the data is generated by a specific statistical model \cite{Statisticalmodeling:}. On the other hand, machine learning methods consider the data generation process as a black box and try to learn from the data only. The machine learning methods are based on the implicit assumption that the underlying data generation process is not relevant as long as the machine learning methods are able to produce accurate predictions \cite{ComparisonStatisticalAndMachineLeanringMethodsMultiStepForecastHydrological}. Thus, the machine learning methods rely on data modelling. \citea{Statisticalmodeling:} regards these two approaches as two different cultures and recommends to use the machine learning approach. There is an ongoing debate about which of these methods perform better. In this paper, we want to evaluate each of them quantitatively to provide more clarity on this subject.

%------------------------------------------------------- SECTION ANOMALY DETECTION APPROACHES FOR TIME SERIES
\section{Selected Anomaly detection approaches for time series}
\label{chapter:chapterAnomalyDetectionApproache}

In this section, various anomaly detection methods are introduced. We divide the approaches in three categories: statistical approaches, classical machine learning approaches and anomaly detection methods using neural networks.

\subsection{Anomaly detection using statistical approaches}
As statistical approaches, we have selected some well-researched regressive models such as AR, MA, ARMA, ARIMA and some of the models that have worked well in the Makridakis competitions (also known as M-Competitions), and also some recently published papers. Although M-Competitions compare statistical forecasting methods, the anomaly detection methods on time-series is closely linked to the forecasting approaches. In this regard, they provide a good reference for effective statistical algorithms in time-series analysis.

\begin{enumerate}
    \item \textbf{Autoregressive Model (AR)}\\
        One of the most basic stochastical models for univariate time-series is the Autoregressive model (AR). AR is a linear model where current value $X_t$ of the stochastic process (dependent variable) is based one a finite set of previous values (independent variables) of length $p$ and an error value $\epsilon$:
        \begin{equation}\label{Equation_AR}
          X_t = \sum_{i=1}^p a_i\cdot X_{t-i} + c + \epsilon_t  
        \end{equation}
        The AR model in Equation \ref{Equation_AR} with a preceding window length of $p$ is also called AR process of order $p$ or AR(p). The error values $\epsilon_t$ are considered to be uncorrelated and have a constant mean of zero and constant variance $\sigma$. In this model, $\epsilon$ is used to determine the anomaly score. \\
        The values of the coefficients $a_1,...,a_p,c$ can be approximated by using the training data and solving the corresponding linear equations with least-squared regression. After that,  $\epsilon_t$ for each $X_t$ can be computed, which represents the anomaly score. Hence, the anomaly score is equal to the difference between the forecasted value and observed one \cite{AnomalyDetectionSymbolicSequencesTimeSeriesData}.\\
        AR models assume that the data is stationary. Thus, it is important to analyse the data and transform it if necessary.
    \item \textbf{Moving Average Model (MA)} \\
        While the AR model considers $X_t$ as a linear transformation of the last $p$ observations of a time-series $\{x_t,x_{t-1},...,x_{t-p}\}$, the \textit{moving average Model (MA)} considers the current observation $X_t$ as a linear combination of the last $q$ prediction errors $\{\epsilon_t,\epsilon_{t-1},...,\epsilon_{t-q}\}$:
        \begin{equation}\label{Equation_MA}
            X_t = \sum_{i=1}^q a_i\cdot \epsilon_{t-i} + \mu + \epsilon_t  
        \end{equation}
        The MA model in Equation \ref{Equation_MA} with a preceding window of length $q$ is also called MA process of order $q$ or MA(q). 
        $\mu$ is the mean of the time-series and the coefficients $\{a_0,...,a_q\}$ are learned from the data. In contrast to AR, learning the coefficients in MA is more complicated. While in the AR model preceding values $\{x_t,x_{t-1},...,x_{t-p}\}$ are known, in the MA model the values of $\{\epsilon_t,\epsilon_{t-1},...,\epsilon_{t-q}\}$ are unknown at the beginning. The errors are known after the model is fitted. Thus, they are optimized sequentially. Therefore, a closed solution for the MA models does not exist and an iterative
        non-linear estimation algorithm is used to solve MA models \cite{Rousseeuw:1987:RRO:40031}. 
        After fitting the model, we use the deviation to detect anomalies like in the AR model.
        
    \item \textbf{Autoregressive Moving Average Model (ARMA)}\\
    Another model is the combination of AR and MA, which is often used for univariate time-series in practise. A time-series of the ARMA(p,q) model is dependent on last $p$ observations and $q$ errors:
    \begin{equation}\label{EquationARMA}
        X_t = \sum_{i=1}^p a_i\cdot X_{t-i} +  \sum_{i=1}^q b_i\cdot \epsilon_{t-i} + \epsilon_t 
    \end{equation}    
    $\{X_T\}$ is an ARMA(p,q) process if $\{X_T\}$ is stationary.\\ ARMA models use less variables in practice compared to AR and MA. However, the main challenge is to select appropriate values for $p$ and $q$. The bigger these two values are, the more likely is it that the model overfits, resulting in too many false negatives in the anomaly detection process. On the other side, if they are chosen too small, the model will underfit and too many false positives will arise, i.e., data points are detected as anomalies although they are not. In both cases, the model is not able to detect the anomalies correctly. \\ 
    There are several ways to fit the model and find appropriate values for $p$ and $q$:
    \begin{enumerate}
        \item \textbf{Using correlograms}: First of all the data must be transformed, if necessary, to become stationary. Each ARMA model has its own specific autocorrelation and partial autocorrelation graphics, which can be visualized in a correlogram. The same is true for AR and MA. Thus, the autocorrelation function (ACF) and partially autocorrelation function (PACF) of the time-series is computed. Then, it will be discovered which $p$ and which $q$ of the ACF and PACF correlogram of ARMA(p,q) are similar to ours. This is an iterative process where different values of $p$ and $q$ are evaluated.
        \item \textbf{Leave-One-Out Cross-Validation}: Another proposed way is using the observed data to minimize the error value by assigning different combinations of $p$ and $q$ using leave-one-out cross-validation \cite{OutlierAnalysisAggarwal2016OA3086742}.
        \item \textbf{Box-Jenkins Method} \cite{BoxJenkinsTimeSeriesAnalysisForecastingandControl}: The Box-Jenkins method which was introduced by George Box and Gwilym Jenkins proposes an iterative method:
        \begin{itemize}
        \item[1.] Identification: Use the data information to select the best model that represents the data by setting the $p$ and $q$ values. Additionally, evaluate whether the data is stationary and transform it if this is not the case. For this purpose, the ACF and PACF plots are helpful. 
        \item[2.] Estimation: The model is fitted to the data so that the parameters $a_i$ and $b_i$ can be estimated. 
        \item[3.] Diagnostic checking: The fitted model is checked with the data to evaluate its performance and if any inadequacies are witnessed. If the result is inadequate, we return to step 1. 
        \end{itemize}
    \end{enumerate}{}
    These approaches are not only used for the ARMA model, but are general methods for all statistical approaches.
    \item \textbf{ARIMA Model}\\
    One of the main problems with datasets is the fact that they can be non-stationary. Stationarity is a precondition for models like ARMA. The ARIMA model is a generalization of the ARMA model. In addition to the $p$ and $q$ parameter, it is also defined by a $d$ parameter which defines the number of times the time-series is differenced. For $d=1$, the time-series $\{x_0,...,x_T\}$ is differenced as follows:
    \begin{equation}\label{EquationDifferencing}
        X_i' = X_i - X_{i-1}, \forall i\in \{1,...,T\}
    \end{equation}
    The effect of differencing is best shown through plotting some data. Figure \ref{fig:StockValueNoDiff} shows the stock data from Section \ref{SectionTimeSeries} for the years 2005 to 2008. The data shows a positive trend. Therefore, we do not have stationary data here. However, Figure \ref{fig:StockWithDiff} shows the daily changes of the same stock over the three years, and the data is stationary now:
    
    \begin{figure}[H]
     \centering
     \captionsetup{justification=centering,margin=2cm}
     \includegraphics[scale=0.5]{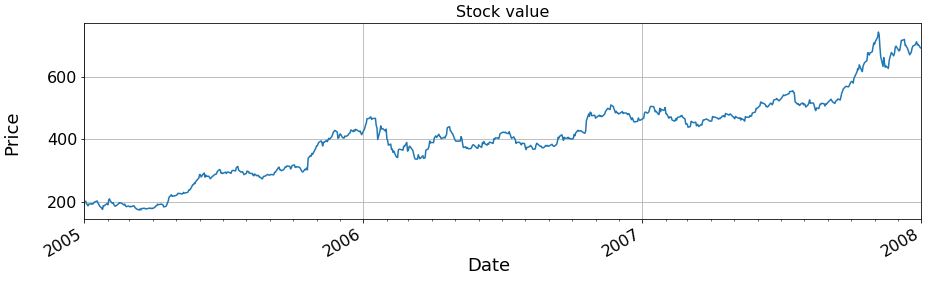}
     \caption{Stock value from 2005 to 2008}
     \label{fig:StockValueNoDiff}
    \end{figure}
    
    \begin{figure}[H]
     \centering
     \captionsetup{justification=centering,margin=2cm}
     \includegraphics[scale=0.5]{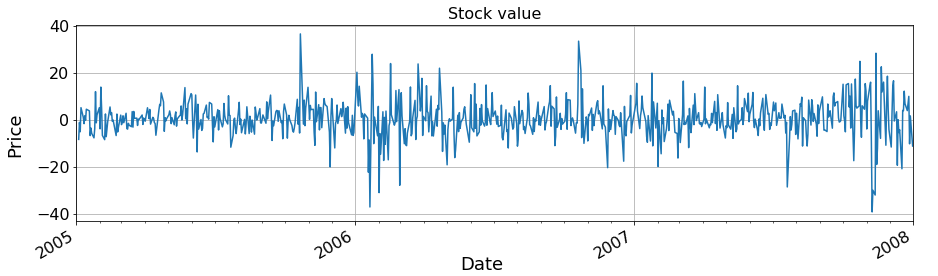}
     \caption{Value changes of the stock value (via differencing method)}
     \label{fig:StockWithDiff}
    \end{figure}

    Figure \ref{fig:StockWithDiff} plots the differences between the consecutive data points of Figure \ref{fig:StockValueNoDiff}.\\
    Differencing removes a trend in the time-series resulting in a constant mean. 
    If the trend is non-linear, differencing must be done several times, thus, $d>1$.\\
    Differencing is also used to remove seasons. The \textit{Seasonal Differencing} is as follows:
    \begin{equation}\label{EquationDifferencingSeasonal}
        X_t' = X_t - X_{t-n} \text{ where n is the duration of the season}
    \end{equation}{}
    After fitting the ARIMA model, anomalies are detected by evaluating the deviation of the predicted point to the  observed one. 
    
    \item \textbf{Simple Exponential Smoothing (SES)} \cite{ExponentialSmoothingBrown}\\
    While in the previous models the prediction  is a linear optimization problem, SES uses a non-linear approach by taking the previous time-series data to predict, assigning exponential weights to the observations:
    \begin{equation}\label{EquationSES}
        \begin{split}
          X_{t+1} = & \alpha X_{t} + \alpha(1-\alpha)X_{t-1} + \alpha(1-\alpha)^2X_{t-2}+ ... + \alpha(1-\alpha)^NX_{t-N}\\ 
          &\text{ where }\alpha\in[0,1]
        \end{split}
    \end{equation}{}
    Thus, $X_{t+1}$ is a weighted combination of the previous data points. The parameter $\alpha$ defines the rate at which the weights decrease, which is exponential. Therefore, it is called \textit{Exponential Smoothing}. The smaller $\alpha$ is, the more weight is given to data points that are more distant. 
    %This is listed in the following table:
    %\begin{figure}[H]
    % \centering
    % \captionsetup{justification=centering,margin=2cm}
    % \includegraphics[scale=0.5]{images/SASAlphaTable.png}
    % \caption{SAS Coefficients for $X_t$ to $X_{t-6}$ for different $\alpha$ %values }
    % \label{fig:SASCoefficient}
    %\end{figure}

    \item \textbf{Double and Triple Exponential Smoothing (DES, TES)} \cite{HYNDMAN2002439}\\
    SES assumes that the data is stationary. SES can be extended to also handle non-stationary data, which is called Double Exponential Smoothing. Here an additional parameter $\beta$ is introduced to smooth the trend in the series. If the data also contains seasonality, Triple Exponential Smoothing is used. This extension also contains a parameter $\gamma$ to control the effect of seasonality. 
    
    \item \textbf{Time-series Outlier Detection using Prediction Confidence Interval (PCI)} \cite{TimeseriesOutlierDetectionPCI}\\
    This approach uses a sequence of previous data which are weighted non-linear to forecast the next data point. Then, by using the threshold, they classify a data point as anomaly or normal. \\
    Thus, to calculate $X_t$, it uses a window of past observed points of the series:
    \begin{equation}
        X_t = \frac{\sum_{j=1}^{2k} w_{t-j}X_{t-j} }{\sum_{j=1}^{2k} X_{t-j}}
    \end{equation}
    where $w_{t-j}$ is the weight for $X_{t-j}$ and it is proportional to the inverse of the distance between $X_t$ and $X_{t-j}$. This gives temporal closer points $X_t$ more weight. 
    If the anomaly detection is done offline, a two sided window can be computed:
    \begin{equation}
        X_t = \frac{\sum_{j=1}^{k} w_{t-j}X_{t-j} + \sum_{j=1}^{k} w_{t+j}X_{t+j} }{\sum_{j=1}^{k} X_{t-j} + \sum_{j=1}^{k} X_{t+j}}
    \end{equation}
    Then, the approach computes an upper and lower bound for the anomaly detection:
    \begin{equation}
        PCI = X_t \pm t_{\alpha,2k-1} \cdot s\sqrt{1+\frac{1}{2k}}
    \end{equation}
    Here, the factor $t_{\alpha,2k-1}$ is the $p$-th percentile of a Student’s $t$-distribution with $2k-1$ degrees of freedom, $s$ is the standard deviation of the model residual, and $k$ is the window size used to calculate $s$. If $X_t$ is outside the boundaries, it is marked as an anomaly.\\
    Thus, this method has some hyperparameters: $\alpha$ to calculate the plausible range of PCI and $k$ as the window size. Here, the analyst faces again the challenge to overcome overfitting and underfitting by adjusting these parameters correctly.\\
    The authors used this method for hydrological time-series data. In their corresponding experiments, they recommend an $\alpha$ value from the interval $[0.85,0.99]$ and $k$ value from the interval $[3,15]$\\
    This method is a simplification of the previous methods as the coefficients are not fitted by the model like AR, MA, ARMA or other autoregression approaches. It does also not use exponential weights like the ES methods. However, it was included in the evaluation of this paper, because it is a more recent approach that was published in 2014. Nevertheless, it is debatable if this method is statistical or a ML method. 

\end{enumerate}{}

\subsection{Anomaly detection using classical machine learning approaches}\label{section_ML}
Machine learning algorithms try to detect anomalies in time-series datasets without assuming a specific generative model. They are based on the fact that it is not necessary to know the underlying process of the data, to be able to make time-series prediction and time-series anomaly detection. Therefore, these methods are well advanced outside the field of statistics \citepa{Statisticalmodeling:}. Many researchers argue that a theoretical foundation of a model can be neglected, if the method performs effectively in practise \citepa{CriteriaClassifyingForecastingMethods}. In this context, we introducte section several univariate anomaly detection methods in this section, which are based on classical machine learning algorithms. Later, the performance of these algorithms are compared to the statistical approaches introduced so far, and the deep learning approaches in Section \ref{Section:AnomalyDetectionNeuralNetwork}.

\begin{enumerate}
    \item \textbf{ K-Means Clustering \--- Subsequence Time-Series Clustering (STSC)}\\
    One of the clustering algorithms for anomaly detection is using K-Means clustering \cite{KMeans}. This method is also called \textit{Subsequence time-series Clustering (STSC)} \cite{WhySubsequenceTimeSeriesClusteringProduceSineWave}. To use K-Means as an anomaly detection method for time-series data, a sliding windows approach is used \cite{FaultDetectionMiningAssociationRulesHouseKeepingData,PracticalMachineLearningNewLookAnomalyDetection}. This implies that given a time-series $\{X_N\} = (x_1, x_2, ..., x_N)$ and window length $w$ and a slide length $\gamma$, the time-series $\{X_N\}$ results in a set of sub sequences $\mathcal{S}\subseteq\mathbb{R}^{(N-w)\times w}$:
    \begin{equation}
        \mathcal{S} = \{(x_0,x_1,...,x_w)^T, (x_{0+\gamma},x_{1+\gamma},...,x_{w+\gamma})^T,...,(x_{N-w},x_{N-w+1},...,x_{N})^T\}
    \end{equation}{}
    After defining the desired number of clusters $k$, the k-Means algorithm is executed on the dataset $\mathcal{S}$ until it converges, resulting in $k$ centroids \cite{RefiningInitialPointsKMeansClustering}. The centroids are the mean of the vectors in the specific cluster. The set of $k$ centroids shape the set $\mathcal{C}$.\\
    To detect anomalies, the distance of each subsequence $s \in \mathcal{S}$ to its nearest centroid is computed, which results in the sequence $\mathcal{E}$:
    \begin{equation}
        \mathcal{E} = (e_0, e_1, ..., e_{|\mathcal{S}|})
    \end{equation}{}
    where $e_i$ for $i \in \{0,...,|\mathcal{S}|$\} is:
    \begin{equation}
       e_i = \min_{\forall c \in \mathcal{C}}(d(s_i-c))
    \end{equation}{}
    where $d$ is a distance function. Usually, the euclidean distance is used for univariate data. \\
    Thus, the sequence $\mathcal{E}$ represents the error value of each sliding window. By defining a threshold $\delta \in \mathbb{R}$, a window $s_i \in \mathcal{S}$ is an anomaly if the corresponding error value $e_i > \delta$.\\
    The main challenge of this approach is specifying an appropriate value $k$. The complexity of this method is O($kNrw$) where $k$ is the number of clusters, $r$ the number of iterations until convergence, $N$ the number of objects (here $N=|\mathcal{S}|$) and $w$ the length of the sliding window \cite{ClusteringTimeSeriesSubsequencesMeaningless}.\\
    \textbf{\textit{Note:}}
    \citea{ClusteringTimeSeriesSubsequencesMeaningless} have demonstrated in their work that using sub-sequences of time-series for clustering algorithms is meaningless. They showed, that the cluster centers found for several runs of the K-means algorithm on the same dataset are not significantly more similar to each other than the cluster centers of a random walk dataset. That means that after being asked to present the centroids on a dataset, they could just present the centroids of a random walk and nobody would be able to distinguish between them \cite{ClusteringTimeSeriesSubsequencesMeaningless}.
    They also tried other algorithms like hierarchical clustering, which is a deterministic approach compared to K-means, but received the same result. The same was proved on several datasets which furthermore confirmed their claim that using sub-sequences of the time-series data for clustering techniques is meaningless. They also tried different distance measures like Manhattan, $L_\infty$ and Mahalanobis distance. Furthermore, by using K-Means with $k = 3$ and $w = 128$ on the famous Cylinder, Bell and Funnell (CBF) dataset, they showed that the resulting centroids are sinus waves, which are totally different to the instances in the CBF dataset. Several authors tried to analyse this behavior mathematically \cite{WhySubsequenceTimeSeriesClusteringProduceSineWave,TheoreticalAnalysisSubsequenceTimeSeriesClusteringFrequencyAnalysisViewpoint,AnalysisSubsequenceTimeSeriesClusteringBasedMovingAverage}, and there have been a lot of attempts to solve these problem, or at least to show time-series patterns that would work with STSC \cite{MakingSubsequenceTimeSeriesClusteringMeaningful,TimeseriesEpenthesisClusteringTimeSeriesStreamsRequreisIgnoringSomeData}. But the problems remain generally unsolved \cite{ReviewSubsequenceTimeSeriesClustering}.\\
    We will use STSC in our evaluation as it still is one of the basic clustering approaches and serves as a comparing artifact. 
    \item \textbf{Density-Based Spatial Clustering of Applications with Noise (DBSCAN)}\\
    Another anomaly detection method based on clustering is Density-Based Spatial Clustering of Application with Noise algorithm (DBSCAN) \cite{DensityBasedAlgorithmDiscoveryClusters}. In comparison to other clustering methods like STSC or CBLOF \cite{CBLOF}, it also analyses the density in the data. \\
    The method classifies the data points into three different categories:
    \begin{itemize}
        \item Core points
        \item Border points
        \item Anomalies
    \end{itemize}
    To classify the points, the user has to specify two parameters: $\epsilon$ and $\mu$ where $\epsilon$ is the distance to declare the neighbors of the analysed point and $\mu$ is the minimum number of points each normal cluster has to have. \\
    To classify a point, first the $\epsilon$-neighbors of each point have to be determined. Thus, for the dataset $\mathcal{D}$ where $\mathcal{D} =\{x_i |x_i\in \mathbb{R},i\in\{1,...,n\}\}$, the $\epsilon$-neighbors of $x_i$ is:
    \begin{Definition}
        $\epsilon-neighbors(x_i) = \{x_j|  x_j\in\mathcal{D}: \phi(x_i,x_j)\leq \epsilon, x_i\neq x_j \} $
    \end{Definition}
    where $\phi$ is a distance function.
    
    Then, a point $x_i \in \mathcal{D}$ is a \textit{Core Point} if :

    \begin{Definition}
        CorePoint($x_i$)$=true \Leftrightarrow \epsilon-neighbors(x_i) \geq \mu $
    \end{Definition}
    
    Border points are declared as follows:
    \begin{Definition}
        $BorderPoint(x_i)=true \Leftrightarrow \exists x_j \in \mathcal{D}: x_j \neq x_i \wedge x_j\in\epsilon-neighbors(x_i)\wedge CorePoint(x_j)=1$
    \end{Definition}
    
    It is also possible to set a threshold $\delta$ for border points, such that it should have more than $\delta$ CorePoints as neighbors. 
    
    Finally, anomalies are defined as follows:
    
    \begin{Definition}
        $Anomaly(x_i)=true \Leftrightarrow CorePoint(x_i) = false \wedge BoarderPoint(x_i) = false$
    \end{Definition}
    %\todo[inline]{bin nicht  sicher, ob der leftrightarrow bei diesen ganzen definitionen das richtige symbol ist - A: leftrightarrow ist das Latexsymbol für if Genau dann wenn und hier gilt $x_i$ ist eine Anomaly genau dann wenn die zwei Bedingungen gelten.}
    
    \citea{DBSCANTimeseries} have used DBSCAN for anomaly detection on a univariate time-series dataset, which contains the daily average temperature observations for 33 years. They first split the dataset into sequences containing the data of a month. Then, the data is normalized using the mean and variance of the data's sequence. After that, DBSCAN is run on each sequence and the anomalies are detected as described before. The main challenge is to select appropriate values for the parameters $\epsilon$-Distance and $\mu$ as the minimum number of points in each cluster.

    % checked until here
    
    \item \textbf{Local Outlier Factor (LOF)}\\
    Another popular clustering algorithm is Local Outlier Factor (LOF) clustering. In contrast to DBSCAN, it is not based on density, but finding the nearest neighbors (K-NN) \cite{ComparativeAnalysisDensityBasedOutlierDetection}, while also focusing on local outliers. LOF was initially designed to detect anomalies on spatial data \cite{LOF}. But \citea{EventDetectionMarineTimeSeriesDataLOF} extended the approach to use it also for time-series data. \\
    Let $\mathcal{D}$ be a dataset and $x \in \mathcal{D}$. To calculate the LOF value of a data point $x$, the following steps have to be performed:
    % \todo{Hast du hier absichtlich Punkt (a) geändert? Ich habe die alte Version kommentiert hinzugefügt, falls du sie mal wieder testen willst -- Ne war keine Absicht.}
    % \begin{enumerate}
    %     \item Compute the k-distance $\delta_k$ of $x$:\\
    %     let $k \in \mathbb{N}_+$ and $\phi$ be a distance function, then $\delta_k$ = $k$-distance of $x$ if:
        
    %   $ \phi(x,y) = \delta_k, \text{ where } x\in \mathcal{D}\text{ and } y \text{ is the $k$-nearest neighbor to }x$.
       
    %   Then, the $k$-distance neighborhood of $x$ is as follows:
       
    %   $N_{k-distance(x))}(x) = \{y| y\in \mathcal{D} , \phi(x,y)\leq\delta_k\}$
       
    %   And the reachability distance RD of $x$ is defined as:
       
    %   $RD_k(x,y) = max\{k-distance(y),\phi(x,y)\}$
       
    %   \item The local reachability density (LRD) of $x$ is computed:
       
    %   $LRD_k(x) = 1/ \left(\frac{\sum_{y\in N_{k-distance(x)}} RD_k(x,y)}{|N_{k-distance}(x)|}\right)$
       
    %   \item Finally, the local outlier factor of $x$ can be computed:
       
    %   $LOF(x) = \frac{\sum_{y\in N_{k-distance(x)}}\frac{LRD_k(y)}{LRD_k(x)}}{|N_{k-distance}(x)|}$
    % \end{enumerate}{} 
    % ALTE VERSION für (a)
    \begin{enumerate}
        \item Compute the k-distance $\delta_k$ of $x$:\\
        let $k \in \mathbb{N}_+$ and $\phi$ a distance function, then $\delta_k$ = k-distance of $x$ if:
      $$ \phi(x,y) = \delta_k, \text{ where } x\in \mathcal{D}\text{ and } y \text{ is the k-th neigherst neighbor to }x$$
      Then k-distance neighborhood of $x$ is the follows:
      $$N_{k-distance(x))}(x) = \{y| y\in \mathcal{D} , \phi(x,y)\leq\delta_k\}$$
      And the reachablity distance RD of $x$ is defined as:
      $$RD_k(x,y) = max\{k-distance(y),\phi(x,y)\}$$
      \item Then the local reachability density (LRD) of $x$ is computed:
      $$LRD_k(x) = 1/ \left(\frac{\sum_{y\in N_{k-distance(x)}} RD_k(x,y)}{|N_{k-distance}(x)|}\right)$$
      \item Finally, the local outlier factor of $x$ can be computed:
      $$LOF(x) = \frac{\sum_{y\in N_{k-distance(x)}}\frac{LRD_k(y)}{LRD_k(x)}}{|N_{k-distance}(x)|}$$
    \end{enumerate}{}

    \citea{EventDetectionMarineTimeSeriesDataLOF} used a sliding window with length $w$ to classify a $w$-long sequence as an anomaly. Thus, given a time-series ${X_t}$, it is split into a training set $A$ and test set $B$. Using the window length $w$, each set is transformed as follows:
    \begin{equation}
        A(w,t)\subseteq P(\{X_T\}) = \{(x_i,...,x_{i+w})|i\in\{1,...,t-w\},t\leq |\{X_T\}|\}
    \end{equation}{}
    $A(w,t)$ is a set of time-series of length $w$.
    \begin{equation}
        B(w,t)\subseteq \{X_T\} = \{x_{t-\frac{w}{2}},...,x_{t+\frac{w}{2}}\}
    \end{equation}
    And $B(w,t)$ is the sequence we want to analyse. \\
    To determine if $B(w,t)$ is an anomaly or not, we compute the anomaly score $\phi$ of $B(w,t)$ and if $\phi > \delta$ where $\delta \in \mathbb{R}$ is a threshold then we mark $B(w,t)$ as an anomaly.
    \\
    The anomaly score $\phi$ of $B(w,t)$ specifies, how much $B(w,t)$ is different than the sets in $A(w,t)$. This is done by computing the LOF value of $\{\{B(w,t)\}\cup A(w,t)\}$. If $LOF(B(w,t))>\delta$, then the sequence is marked as an anomaly.
    \\
    The main challenges of this approach are the following items:
    \begin{itemize}
        \item Determine an appropriate value $k$ for the k-nearest neighbors to compute the LOF value. In general, it is considered that prior knowledge is available to determine an appropriate value for $k$. The authors of the paper suggest using an ensemble strategy to compute $k$ \cite{ComparativeEvaluationUnsupervisedAnomalyDetectionMultivariateData}.
        \item Concatenating the time-series into a vector and computing the distance to other vectors removes the ordered information of a time-series. Here, temporal data is converted into spatial data, where each dimension is equally important. But in time-series the ordered sequence contains important information, which is used in some statistical approaches like Exponential Smoothing. 
        \item Determine an appropriate distance function $\phi$. While  \citeauthor{ComparativeEvaluationUnsupervisedAnomalyDetectionMultivariateData} recommend using the Euclidean distance, there have been many cases where the Euclidean distance is not suitable, especially to also consider the relationship between variables in a multidimensional space.
        \item The LOF value only relies on the direct neighbors, which also makes it more appropriate to detect local anomalies. 
        \item Another drawback of the LOF algorithm is the complexity of $O(n^2)$ as compared to DBSCAN with a complexity of $O(n\cdot log~n)$.
    \end{itemize}{}

    \item \textbf{Isolation Forest}\\
    One of the machine learning approaches to detect anomalies in time-series is isolation forest using a sliding window. Isolation forest, also known as iForest, was introduced by \citea{IsolationForest}. It builds an ensemble of Isolation Trees (\textit{iTrees}), which are binary trees isolating data points. As anomalies are more likely to be isolated than non-anomalous points, it is more likely that they are closer to the root of an iTree \cite{IsolationForestAnomalyDetection}. Figure~\ref{fig:IsolationForest} shows how an iTree is generated on  a sample data structure. An anomaly is detected after two partitions while the first normal point is detected after the fourth partition:
    \begin{figure}[H]
    \centering
        \captionsetup{justification=centering,margin=2cm}
        \includegraphics[scale=0.4]{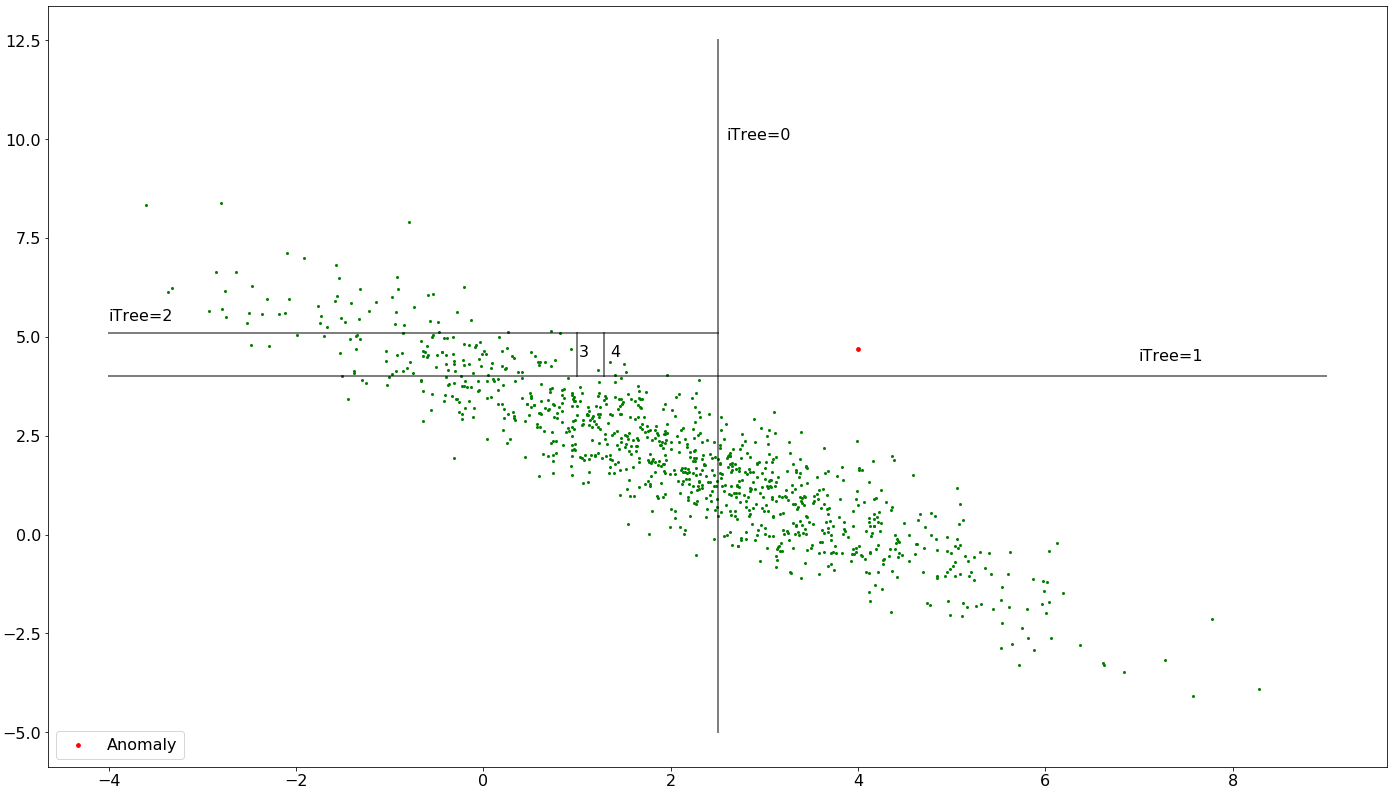}
        \caption{Isolation Forest: The anomaly is isolated by the random generated tree after two partitions}
    \label{fig:IsolationForest}
    \end{figure}
    Thus, this method considers points with shorter path lengths as candidates that are highly likely to be anomalies. \\
    The anomaly detection process using Isolation Forests is performed generally in two steps:
    \begin{enumerate}
        \item Training: Create $n$ iTrees for the given training set.
        \item Evaluation: Pass the test instance through the isolation trees to determine the anomaly score. 
    \end{enumerate}{}
    There are some methods that have extended iForest to detect anomalies on time-series. One main approach to detect anomalies in univariate data is to analyse the dataset in sequences defined by the window length $w$. Thus, let $\{X_T\} = (x_1, x_2, ..., x_T)$ be a univariate time-series, $w$ the window length and $W \subseteq  \mathbb{R}^{p\times w}$. Then:
    
    \begin{equation}
        \begin{split}
        {W} := &(W_1, W_2, ..., W_p)\\
            = & ((x_1, ..., x_w)^T,(x_2,..., x_{w+1})^T,...,(x_p, ..., x_{w+p-1})^T)
        \end{split}
    \end{equation}{}
    After that, the anomaly score on each sequence is computed, which is proportional to the average path length of an instance. Using supervised learning, the threshold can be computed on the training set and used for the test set later. \\
    \citea{AnomalyDetectionIsolationForestStreamingDataSlidingWindow} have extended the concept to compute the anomaly score $S(x,p)$, where $x$ is the data point and $w$ is size of the window:
    \begin{equation}
        \begin{array}{l}
        S(x,w) = 2^{-\frac{E(h(x))}{c(w)}}\\
        E(h(x)) = \frac{1}{L} \sum_{i=1}^L h_i(x)
       \end{array}
   \end{equation}{}
    where $h_i(x)$ denotes the length of the i-th iTree, $E(h(x))$ the average of $h(x)$ from a collection of iTrees and $c(p)$ is the average of $h(x)$ given $w$ and $L$, the number of iTrees.\\ 
    The main challenges of the isolation forest algorithm are the following parameters:
    \begin{itemize}
        \item Window length $w$: If the length is too short, then there will not be enough data to construct an appropriate model. On the other hand, if the length is too long, older and sometimes less relevant data will be considered as much as more recent data points. \citea{AnomalyDetectionIsolationForestStreamingDataSlidingWindow} have shown in their experiment results that fixed sliding windows for different datasets result in bad performance.
        
        \item Number of iTrees in the iForest: The higher the number of iTrees, the closer the average value is to the expected value. The downside is that a higher number of iTrees will increase the computation time \cite{IsolationForestAnomalyDetection}. For $w$ as the window length and $L$ as the number of iTrees, iForest has a time complexity of $O(L\cdot w^2)$ and a space complexity of $O(L\cdot w)$.
        
        \item Contamination: Many implementations of iForest, e.g. the implementation in \textit{sklearn}, have a contamination parameter where the proportion of anomalies in the  dataset is set. This also marks the threshold for the anomaly. Improper assignment of this parameter could result in a higher rate of false positive or false negatives.
    \end{itemize}{}

    \item \textbf{One-Class Support Vector Machines (OC-SVM)}\\
    The original support vector machine algorithm was invented as a linear supervised approach by \citea{SVMVapnikChervonenkis} in 1963. Boser et al. extended the algorithm by introducing the kernel trick, which made SVM capable of making non-linear classification. After that, a new approach to detect novelties using SVM was introduced called One-Class SVM (OC-SVM)\cite{SVMNoveltyDetection}. OC-SVM is a semi-supervised approach where the training set consists of only one class: the \textit{normal} data. After the model is fitted on the training set, the test data is classified as being similar to the normal data or not, making it able to detect anomalies. \\
    The original OC-SVM method was able to detect anomalies in a set of vectors and not on time-series. Most papers recommend to project the time-series into a vector set. \citea{TimeseriesNoveltyDetectionOCSVM} propose to unfold the time-series into a phase space using a time-delay embedding process \citepa{GeometryFromTimeSeries}.
    \citea{OneClassSVMAnomalyDetectionCommunicationNetworkperformanceData},\cite{NetworkAnomalyDetectionOneClassSVM} recommend to create windows with length $w$ of the time-series dataset, so that for a given time-series $\{X_T\} = (x_1, x_2, ..., x_T)$, a window length $w$ and $W \subseteq  \mathbb{R}^{p\times w}$, the dataset is first converted into :
    %\todo{citation? - Antwort: Ich habe das geändert, aber in der vorherigen Version war ja auch citation dabei, aber am Ende des Satzes. Das Problem ist, dass es zwei Zitate sind. Ich habe das jetzt so geändert. Unten ist die vorherige Version, falls du sie doch beibehalten willst.}
    % Zhang et al.\todo{citation?} recommend to create windows with length $w$ of the time-series dataset, so that for a given time-series $\{X_T\} = (x_1, x_2, ..., x_T)$, a window length $w$ and $W \subseteq  \mathbb{R}^{p\times w}$, the dataset is first converted into \cite{OneClassSVMAnomalyDetectionCommunicationNetworkperformanceData,NetworkAnomalyDetectionOneClassSVM}:
    \begin{equation}
        \begin{split}
        {W} := & (W_1, W_2, ..., W_p)\\
        = & ((x_1, ..., x_w)^T,(x_2,..., x_{w+1})^T,...,(x_p, ..., x_{w+p-1})^T)            
        \end{split}{}
    \end{equation}{}
    Then, a function $p$ projects the time-series into a two dimensional space:
    \begin{equation}
        \begin{array}{l}
             f: \mathbb{R} \to \mathbb{R}^2\\
             f(X_t) \mapsto 
             \begin{cases*}
            [X_t X_t], &\text{ for } t=1\\
            [X_t X_{t-1}], &\text{ else }
            \end{cases*}
            
        \end{array}{}
    \end{equation}
    
    While in OC-SVM, the result is biased on time-series points with large values, it is recommended that the data is normalized. 
    
    %---------------------------- XGBOOST -------------------------------
    \item \textbf{Extreme Gradient boosting (XGBoost, XGB)}\\
    A machine learning technique which also performed well in the Kaggle and KDDCup competitions is Extreme Gradient boosting (XGBoost). One of the main advantages of XGBoost is its scalability \cite{XGBoost}. XGBoost is derived from the Tree boosting algorithm using the second order method introduced by \citea{AdditiveLogisticRegression}.\\
    Let $D$ be a dataset of n examples with dimensionality $m$: $D=\{(x_i,y_i) | x_i \in \mathbb{R}^m, y\in\mathbb{R}, i\in \{1,...,n\}\}$. Then, tree boosting uses a sequential sequence of tree models to make a prediction $\hat y$ for $x_i$:
    \begin{equation}
        \hat{y_i} = \phi(x_i) = \sum_{k=1}^K f_k(x_i), \text{ where } f_k \in \mathbb{F}
    \end{equation}{}
    where $\mathbb{F}$ is the space of regression trees. The corresponding loss function is:
    \begin{equation}\label{LossFunctionTreeboosting}
        \begin{array}{l}

        \mathcal{L}(\phi) = \sum_il(\hat{y_i},y_i) + \sum_k\Omega(f_k)\\\\
        \text{where } \Omega(f) = \gamma T +\frac{1}{2} \lambda ||w||^2
        \end{array}{}
    \end{equation}{}
    where $T$ is the number of leaves in each tree and $w$ the leaf weights. \\
    The loss function in Equation \ref{LossFunctionTreeboosting} contains functions, which are not possible to optimize with traditional optimization methods in Euclidean space. To work around this obstacle, the model is trained in an additive manner and the taylor approximation of the loss function is used to make it optimizable in the Euclidean space.
    \begin{equation}
        \mathcal{L}^{(t)} \approx \sum_i[l(\hat{y_i}^{t-1},y_i) + g_if_t(x_i) + \frac{1}{2}h_if^2_t(x_i) ] + \Omega(f_t)
    \end{equation}{}
    where $g_i$ is the derivative of the loss: $\partial_{\hat{y}^(t-1)} l(y_i,\hat{y}^{(t-1)})$ and $h_i$ the second derivative: $\partial_{\hat{y}^(t-1)}^2 l(y_i,\hat{y}^{(t-1)})$.\\
    Tianqi Chen et al. provide more details in their main paper about XGBoost \cite{XGBoost}. \\
    Thus, XGBoost is used as a regression model to forecast time-series. To detect anomalies in univariate time-series, we extend the algorithm to compute the error in the prediction. Based on the training data, we are able to compute $\lambda$ percentile of the error distribution and mark it as the threshold $\delta \in \mathbb{R}$ to detect anomalies.

\end{enumerate}{}

\subsection{Anomaly detection using neural networks}\label{Section:AnomalyDetectionNeuralNetwork}
Since neural network have achieved tremendous results in computer vision tasks like object detection, classification and segmentation or similar tasks there have been increasing interest to use them for time-series forecasting and time-series analysis. They are similar to classical machine learning approaches with regard to the fact that they do not presume any knowledge of the underlying data generation process. Their popularity is based on their empirical results.\\
Many researchers have tried to evaluate the performance of neural networks compared to the classical approaches like ARIMA. \citea{ConnectionistApproachTimeSeriesPredictionEmpiricalTest} compared 101 time-series using forward neural networks and the ARIMA model.
\citea{FeedForwaredNeuralNetsTimeSeriesForecasting} also compared neural networks with ARIMA models focusing on 16 time-series with different complexities. Using neural networks for time-series forecasting paved the way for using neural networks to detect anomalies in time-series. In this section, we selected the most popular approaches used in recent years.  
\begin{enumerate}
    %---------------------------- MLP -----------------------
    \item \textbf{ Multiple Layer Perceptron (MLP)}\\
    The most fundamental artificial neural network architecture (ANN) is the Multilayer Perceptron (MLP) \cite{ElementsStatisticalLearningDataMiningInferencePrediction} which is a fully-connected feed-forward neural network. According to  \citea{HyndmanforecastingPrinciplesAndPractice} a neural network used for time-series prediction is a  \textit{Neural Network Autoregression Model (NNAR Model)}. They characterize a NNAR model by the lagged input $p$ and the nodes in the hidden layer $k$: NNAR$(p,k)$. Thus:
    \begin{equation}
        \text{NNAR}(p,0)  \Leftrightarrow  ARIMA(p,0,0)
    \end{equation}{}
    where no seasonal restriction exists. Thus, $p$, which is the lagged input, represents also the window size $w$ of the sliding window used on the time-series. The window size is equal to the number of neurons in the input layer of the MLP (Figure \ref{fig:MLP}):
    \begin{figure}[H]
    \centering
    \captionsetup{justification=centering,margin=2cm}
    \includegraphics[scale=0.5]{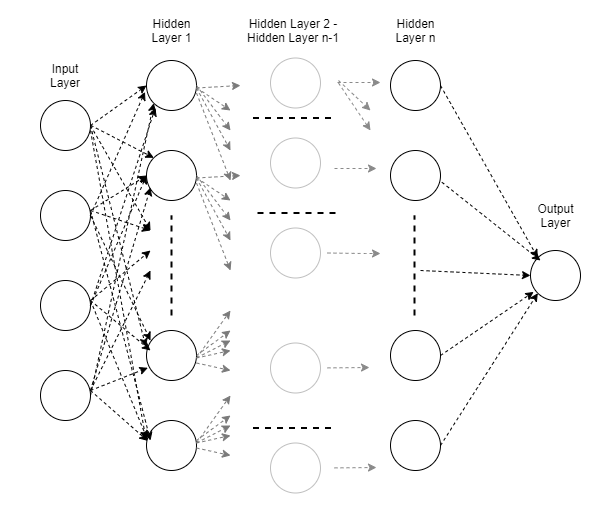}
    \caption{Multilayer Perceptron (MLP)}
    \label{fig:MLP}
    \end{figure}
    While Hyndman et al. use a neural network with one hidden layer and extend the number of neurons in the layer, it is also possible and sometime preferable to increase the number of hidden layers \cite{DoDeepNetsReallyNeedBeDeep}.
    
    \citea{UsingTimeDependenNeuralNetworksEGGClassification} used two different topologies of MLP for time-series classification which is a similar problem to anomaly detection. On the one hand, they implement an MLP using time-series with sliding window and on the other hand, a MLP with finite impulse response filters (FIR-MLP) is used \cite{FIRandIRRSyanpsesNeuralnetworkTimeSeriesModeling}.\\
    In this survey, we will focus on MLP by sliding window on the time-series. The MLP network is used to make predictions. After that the error of the prediction is used to classify a data point as normal or as an anomaly considering the error value proportional to the anomaly score. \\
    Thus, let $\{X_N\}$ be a time-series, $x_i \in \{X_N\}$, $w$ the window length, and $f$ the function of the MLP, then:
    \begin{equation}
        \hat{x}_{t+1} = f(x_{t-w},...,x_t), \forall t \in \{w,...,n\}
    \end{equation}{}
    Hence, the label for a window of time-series $(x_{t-w},...,x_t)$ is the next datapoint of the time series: $x_{t+1}$.\\
    An MLP and all other neural network approaches can also be used to predict more than one timestamp:
    \begin{equation}
        (\hat{x}_{t+1},...,\hat{x}_{t+p_w}) = f(x_{t-w},...,x_t)
    \end{equation}{}
    $p_w$ is the number of timestamps the MLP predicts which is called the \textit{prediction window} or \textit{Forecasting Horizon}\cite{DeepANT_DeepLearningUnsupervisedAnomalyDetectionTimeSeries}. \\
    The prediction of the MLP is then used to detect anomalies. Let $\delta \in \mathbb{R}$ be the anomaly threshold, then $x_{i+1}$ is marked as an anomaly, if:
    \begin{equation}
        f(x_{i-w},...,x_i) - x_{i+1} > \delta
    \end{equation}{}
    The training set of the time-series data can be used to detect a proper value for $\delta$.\\
    One of the main challenges of neural networks is the hyperparameter tuning task. A MLP has a remarkable amount of hyperparameters:
    \begin{enumerate}
        \item Depth of the MLP (Amount of the hidden layers of the network)
        \item Width of the MLP (Amount of nodes in each layer)
        \item Length of the window $w$
        \item Learning rate
        \item Optimization function
    \end{enumerate}{}
    These parameters can be optimized using random search or more advanced techniques \cite{HyperparameterOptimization}.

    %------------------- CNN ----------------------
    \item \textbf{Convolutional Neural Networks (CNN)}\\
    Another artificial neural network approach which is used for anomaly detection in time-series data are deep convolutional neural networks (CNNs). CNNs are mainly used in computer vision for tasks like object detection, classification and segmentation \cite{ImageNetClassificationDeepConvolutionalNeuralNetwork,UNET,ObjectDetectionDeepLearningReview}. In contrast to MLPs, where layers are fully connected, a CNN uses convolutional layers that are partially connected, reducing the amount of parameters, enabling them to go deeper and train faster. CNNs, in contrast to MLP, focus on local patterns in the data. In addition to the convolution layers, CNNs also use pooling layers for regularization to avoid overfitting. One of the pooling operations, which achieves good results regularly \cite{EvaluationPoolingOperationsCNN}
    is the maximum pooling \cite{MaxPoolingDropoutRegularizationCNN}.\\
    In the recent years, there has been increasing interest in using CNNs for time-series analysis. 
    \citea{DeepANT_DeepLearningUnsupervisedAnomalyDetectionTimeSeries} use a CNN architecture, called deep-learning based anomaly detection approach (DeepAnT), to forecast time-series and detect anomalies based on the error of the prediction. \citea{TimeSeriesClassificationMultiChannelsDeepConvolutionNeuralNetworks} use a similar CNN architecture for classification of time-series data, a method that can also be extended to detect anomalies. \\
    Using CNNs for time-series analysis is different to using CNNs for image classification. While the input for image classifying CNNs is 2D, univariate time-series CNNs use 1D input. Therefore, the kernels of the convolution layers are 1D, too.  Figure \ref{fig:DeepAnT} shows the architecture used in DeepAnT:
    \begin{figure}[H]
    \centering
    \captionsetup{justification=centering,margin=2cm}
    \includegraphics[scale=0.5]{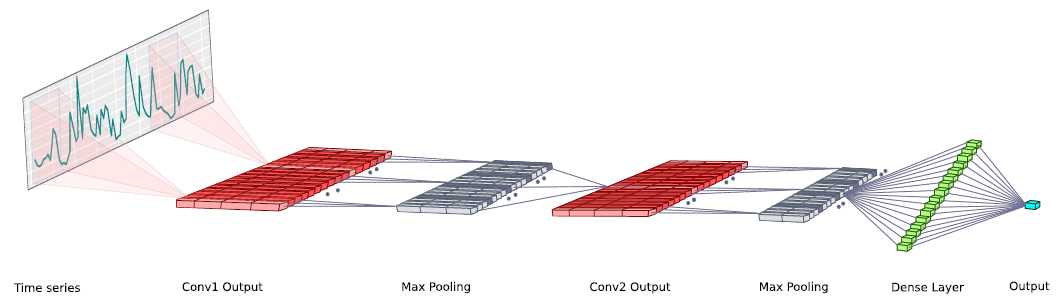}
    \caption{DeepAnT architecture for time-series prediction \cite{DeepANT_DeepLearningUnsupervisedAnomalyDetectionTimeSeries}}
    \label{fig:DeepAnT}
    \end{figure}
    The first layer after the input layer is a 1-dimensional convolution layer followed by a max-pooling layer. As figure \ref{fig:DeepAnT} shows, DeepAnT uses two pairs of convolution and max-pooling layers. However, this could vary based on the dataset. Therefore, one of the major hyperparameters of neural networks in general and CNNs in particular is the architecture of the model. The amount of convolution and max-pooling layers differ in the architectures used in this paper w.r.t. the dataset. We will analyse this in Section \ref{chapter:Experiments}.
    
    After the convolution and max pool layers, a dense layer is used which is fully connected to the output node. If the prediction window is greater than one, the amount of the output nodes will increase accordingly. As an activation function for the convolution layer and the dense layer,  rectified linear units (ReLU) are used \cite{RELU}. \citeauthor{BatchNormalization} suggest another regularization technique called Batch Normalization \cite{BatchNormalization}. Experiments in computer vision showed that Batch Normalization results in a higher learning rate, acts as an alternative for dropout layers and decreases the importance of careful parameter initialization. Therefore, we also implement a CNN architecture using Batch Normalization for univariate anomaly detection to evaluate its performance in anomaly detection. \\
    The CNN model is used to make a prediction in the same way as the MLP model. To detect the anomalies, the same algorithm is used. \\
    Let $\delta \in \mathbb{R}$ be the anomaly threshold, $f$ the function implemented by the CNN model, then $x_{i+1}$ is marked as an anomaly, if:
    \begin{equation}
        f(x_{i-w},...,x_i) - x_{i+1} > \delta
    \end{equation}{}
    In addition to the hyperparameters that MLP also had to handle, CNN expects the following:
    \begin{enumerate}
        \item Architecture of the CNN, i.e., using Batch Normalization, Dropout or Max Pooling layers
        \item Amount of kernels in each convolution layer
        \item The size of the kernel    
        \item Depth of the Convolution Layer
    \end{enumerate}{}
    
    \item \textbf{Residual Neural Network (Resnet)}\\
    An extension of the CNN model, which achieved good results in the last years is the Residual Neural Network (ResNet) architecture. ResNet introduces a new artifact called residual blocks developed by  \citea{DeepResidualLearningImageRecognition}. Residual blocks add the output of a convolutional block with the input of it using a skip connector. A convolutional block consists of several convolution layers, activation layers and regularization artifacts such as Max-Pooling or Batch Normalization layers. \\
    To express residual blocks formally, let $x_i$ be the input and $\phi$ be a convolution block, then the output of a residual block $y$ is as follows:
    \begin{equation}
        y = \phi(x_i) + x_i
    \end{equation}{}
    Usually, an activation function $\psi$ like the RelU activation function is used as well:
    \begin{equation}
        y = \psi(\phi(x_i)+x_i)
    \end{equation}{}
    Figure \ref{fig:ResidualBlock} shows a residual block that is used in this paper:
        \begin{figure}[H]
    \centering
    \captionsetup{justification=centering,margin=2cm}
    \includegraphics[scale=0.7]{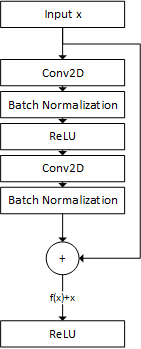}
    \caption{Residual Block used for time-series consisting of convolution block of two convolution layers with one ReLU and two Batch Normalization layers}
    \label{fig:ResidualBlock}
    \end{figure}
    Residual Blocks are used to avoid the vanishing gradient problem, which occurs often in deeper CNNs. \\
    \citea{TimeSeriesClassificationResNet} use ResNet to classify time-series data. They use three residual blocks with 64, 128 and 128 filters. As compared to our residual block in Figure \ref{fig:ResidualBlock}, they use three convolution layers and Batch Normalization layers with ReLU activation function. We tried different amount of residual blocks, which will be explained in detail in Section \ref{chapter:Experiments}. \\
    ResNet is best suited for large amounts of data. Therefore, it could overfit on the time-series data if the size of the data is too limited. But as we have achieved good results with ResNet in computer vision tasks, it would be of interest to evaluate this model on time-series data to detect anomalies, too. 
    
    %----------------------- Wavenet -------------------    
    \item \textbf{WaveNet}\\
    WaveNet was developed by \citea{Wavenet} as a deep generative model to create raw audio waveforms. Especially the ability to approximate the predictive distribution of each audio sample conditioned the previous ones, makes it a proper candidate for time-series forecasting. Thus, WaveNet %, which was constructed to create audio waveforms, 
    is a probabilistic model that tries to approximate the joint probability of a waveform $x = \{x_1, x_2, ..., x_T\}$: 
    \begin{equation}
        p(x) = \prod_{t=1}^T p(x_t|x_1,...,x_{t-1})
    \end{equation}
    which makes audio sample $x$ dependent on all samples before.\\
    To accomplish that, WaveNet uses a specific kind of convolution layers: dilated convolution layers. Normal convolution layers use filters. A filter uses a convolution operation on the data with the same size as the size of the filter. Figure \ref{fig:WaveNetNormalConv} shows a CNN with regular convolution layer:
     \begin{figure}[H]
    \centering
    \captionsetup{justification=centering,margin=2cm}
    \includegraphics[scale=0.4]{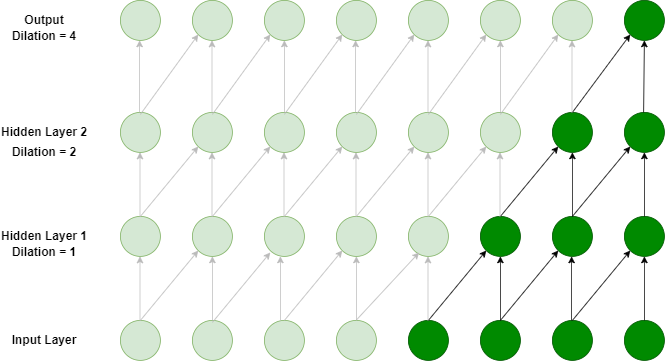}
    \caption{CNN with regular convolution layers with filter size of $2$}
    \label{fig:WaveNetNormalConv}
    \end{figure}   
    In contrast, the dilated convolution layer performs the convolution operation on data bigger than the filter size. This is accomplished by skipping some input values using a skip step. Figure \ref{fig:WaveNetDelationConv} shows a CNN with delation convolution layers:
    
    \begin{figure}[H]
    \centering
    \captionsetup{justification=centering,margin=2cm}
    \includegraphics[scale=0.4]{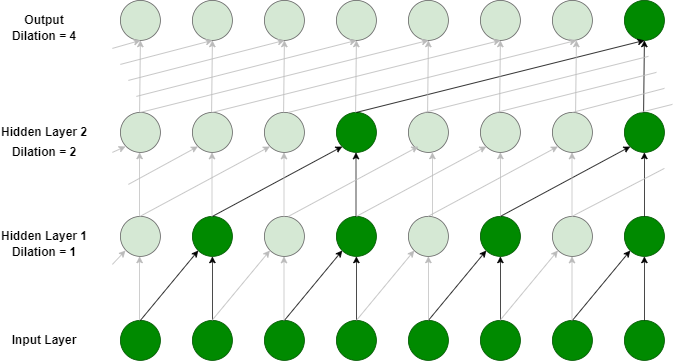}
    \caption{CNN with dilated convolution layers with filter size of $2$ and changing skip steps (delation) = $\{1,2,4\}$}
    \label{fig:WaveNetDelationConv}
    \end{figure}
    One big advantage of the dilation convolution layers is that it will learn long term and short term dependencies while normal convolution layers are designed to extract local patterns. \\
    \citea{ConditionalTimeSeriesForecastingConvolutionNeuralNetworks} used the concept of WaveNet to design a CNN for time-series forecasting. Thus, to predict a timestamp $x_t \in \{X_N\}$, a sequence of timestamps with width $w$ is used as the input for the function $f$ which is expressed by the CNN:
    \begin{equation}
        \hat{x}_t = f((x_{t-w},...x_{t-1})^T)
    \end{equation}{}
    The dilation of the convolution layers increases by a factor of $2$, which is illustrated in Figure \ref{fig:WaveNetDelationConv}. Therefore, the window width $w$ can be much bigger than the value used in normal convolution layers.\\
    In this paper, we will extend this approach to also detect anomalies where we use the same approach we used for the MLP and CNN: \\
    Let $\delta \in \mathbb{R}$ be the anomaly threshold, then $x_{t}$ is marked as an anomaly, if:
    \begin{equation}
        \hat{x}_t - x_{t} > \delta
    \end{equation}{}

    %----------------------- LSTM ------------------------
    \item \textbf{Long Short Term Memory (LSTM) network }\\
    Another ANN, which is designed for sequence data is the LSTM network. LSTM networks belong to the recurrent neural networks (RNN) architectures. In contrast to MLP and CNN, where the data is just flowing forward and therefore also called \textit{feed-forward neural networks}, RNN networks have a feedback connection enabling them to use the output information for the next input of the sequence. Formally, the output of a neuron in a feed forward neural network is as follows:
    \begin{equation}
        y_t = \phi(x_t^T \cdot w_x + b)
    \end{equation}{}
    where $\phi$ is a non-linear activation function. In contrast, the output of a neuron of a recurrent neural network is the following:
    \begin{equation}\label{EquationRNN}
        y_t = \phi(x_t^T \cdot w_x + y_{t-1}^T \cdot w_y + b)
    \end{equation}
    A neuron in a simple RNN is computed as shown in Equation \ref{EquationRNN}. \citea{Hochreiter:LSTM} developed a new version of recurrent cells called Long Short Term Memory (LSTM). Figure \ref{fig:LSTM} shows the structure of an LSTM cell. 
    
    \begin{figure}[H]
    \centering
    \captionsetup{justification=centering,margin=2cm}
    \includegraphics[width=0.9\textwidth]{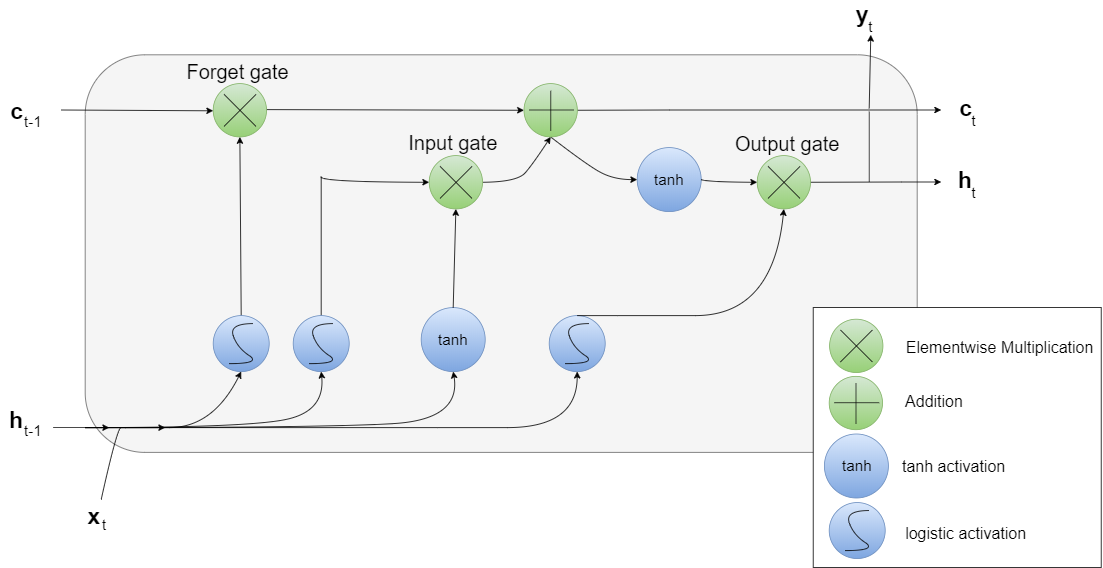}
    \caption{LSTM Cell}
    \label{fig:LSTM}
    \end{figure}
    In contrast to a simple recurrent neuron, LSTM reuses two vectors: $c_t$ and $h_t$. $h_t$ is added with the new data $x_t$ making it a short term memory. On the other side, $c_t$ is multiplied with the new value making it a long term memory. The three gates regulate how much of the data is kept, forgot and delivered to the output. This design aims to recognize important input, and by using addition to the output of the input gate storing it in the long-term state. Additionally, by using the logistic regression and element-wise multiplication, it determines which elements of the long-term memory should be erased. And finally the output gate specifies which part of the new long-term memory is going to be output. \\
    While most neural network architectures like MLP, CNN and simple RNN suffer from the vanishing gradient problem, where the weight updates in the back-propagation step becomes very small, LSTM cells overcome this problem due to its gates, especially the forget gate. \\
    There have been different works using LSTM for univariate and multivariate time-series analysis. It's recurrent manner makes it an appropriate method for sequence data especially time-series. Additionally, most LSTM methods do not use a sequence of timestamp as input for the LSTM model as other approaches like MLP, CNN, ResNet did, as its long and short memory keeps the information of the recent timestamps. Hence, the input of the LSTM consist of just one timestamp, which accelerates the learning process.\\
    \citea{AnomalyDetectionECGTimeSignalsDeepLSTMnetwork} use a LSTM model to predict healthy electrocardiography (ECG) signals. By using the probability distribution of the prediction error, it is able to mark timestamps as normal or anomalous. 
    \\
    
    \begin{figure}[H]
    \centering
    \captionsetup{justification=centering,margin=2cm}
    \includegraphics[scale=0.5]{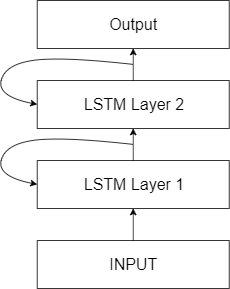}
    \caption{Stacked LSTM: The original LSTM model consisted of just one LSTM layer. Stacked LSTM has multiple LSTM layer}
    \label{fig:StackedLSTM}
    \end{figure}
    \citea{LSTMAnomalyDetectionTimeSeries} use a stacked LSTM (Figure \ref{fig:StackedLSTM}) model consisting of two hidden LSTM layers to predict the next $l$ timestamps. Hence, the prediction window $p_w=l$ where $l>1$. Let again $\{X_T\}$ be a univariate time-series with and $x_i \in \{X_T\}$, then:
    \begin{equation}
        (\hat{x}_{i+1},...,\hat{x}_{i+l}) = f(x_i)
    \end{equation}{}
    \begin{equation}
        \forall i\in \{1,...,t\}, l<i<t-l, \text{ each }x_i\text{ is predicted l-times}
    \end{equation}{}
    Then, the error value of each prediction is computed:
    \begin{equation}
        e^{(i)} = (e_1,...,e_l) = (f(x_{i-l})_l-x_{i+1},f(x_{i-l+1})_{l-1}-x_{i+1},...,f(x_i)_1-x_{i+1})
    \end{equation}{}
    After that, the error vector is used to fit to a multivariate Gaussian distribution $\mathcal{N} = \mathcal{N}(\mu, \Sigma)$. By using Maximum Likelihood Estimation (MLE), the parameters $\mu$ and $\Sigma$ can be computed, which enables it to give any $e^{(i)}$ a likelihood $p^i$. Finally given a anomaly threshold $\delta \in \mathbb{R}$:
    \begin{equation}
        x_i \text{ is an anomaly} \longrightarrow P(e^{(i)})_{e^{(i)}\sim\mathcal{N}(\mu,\Sigma)}>\delta
    \end{equation}{}
    Like the methods we used for the MLP and CNN,  we can use the training set to compute an appropriate value for $\delta$.\\
    
    %---------------------    GRU      ----------------------
    \item \textbf{Gated recurrent unit (GRU)}\\
    In 2014, \citea{LearningPhraseRepresentationGRU} proposed a simplified version of the LSTM cell: Gated recurrent unit (GRU). GRU couples the input and forget gate into one \textit{forget gate}. The state vectors $c$ and $h$ are merged into one vector $h$. Additionally, the output gate is removed and the full state vector is the output at every timestamp. Figure  shows the architecture of GRU:
    \begin{figure}[H]
    \centering
    \captionsetup{justification=centering,margin=2cm}
    \includegraphics[width=0.8\textwidth]{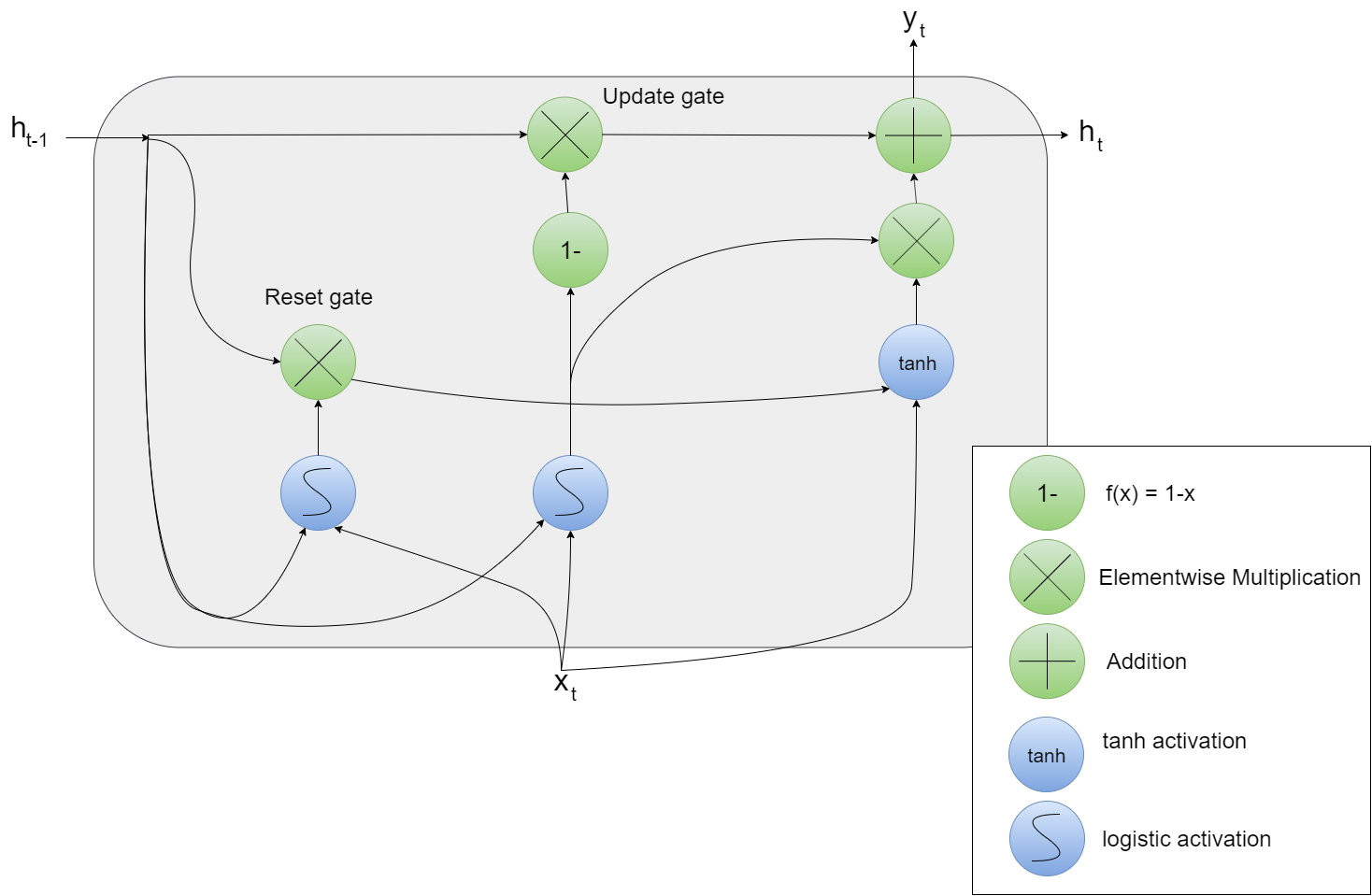}
    \caption{Gated recurrent unit (GRU)}
    \label{fig:GRU}
    \end{figure}
    Research has shown that GRUs perform as well as LSTMs, although requiring less computation due to its simplified structure \cite{LSTMSearchSpaceOdyssey}.\\
    GRUs have also been used for anomaly detection on time-series data. \citea{LocalTrendInconsistencyGRU} used a stacked GRU model to detect anomalies on online time-series data.\\
    In this work, we will also evaluate a GRU model to detect anomalies in univariate time-series. The model will be equal to the LSTM model for anomaly detection for time-series data we explained before. The only difference is that the LSTM cells will be replaced by GRU cells. 
    
    %--------------------- Autoencoder ----------------------
    \item \textbf{Autoencoder}\\
    One method to detect anomalies is to reduce the dimensionality of the data and to project it on a lower space, i.e., latent space, where more correlated variables remain. The main assumption about the distribution of data is the fact that normal and abnormal data are significantly different on this space, which the definition of anomalies (Definition \ref{def:DefinitionAnomaly}) implies. Then, projecting back to the original space will show significant differences in some data points, which represent the anomalous data instances. This makes the autoencoder appropriate for anomaly detection. \\
    Autoencoders belong to the family of feed-forward neural networks, and are optimized to output the same information that was inserted in the network. The challenge is that the first half of the hidden layers reduces the dimension of the dataset, and the second half increases the dimension back to the original value. These two parts are named accordingly the \textit{Encoding} and \textit{Decoding} part. Formally, let $X$ be the dataset and $\psi$ the decoding function and $\phi$ the encoding function and $f$ the corresponding function of the autoencoder, then:
    \begin{equation}
        \hat{X} = \psi(\phi(X))
    \end{equation}{}
    The optimization function of the autoencoder tries to minimize the deviation between $X$ and $\hat{X}$:
    \begin{equation}
        \min_{\theta_{\psi},\theta_{\phi}}\norm{X-\hat{X}}_2= \min_{\theta_{\psi},\theta_{\phi}}\norm{X-\psi(\phi(X))}_2
    \end{equation}{}
    where $\theta_{\phi}$ and $\theta_{\psi}$ are the weights of the decoding and encoding part. Figure \ref{fig:Autoencoder} shows the concept of an autoencoder graphically:
    \begin{figure}[H]
    \centering
    \captionsetup{justification=centering,margin=2cm}
    \includegraphics[scale=0.4]{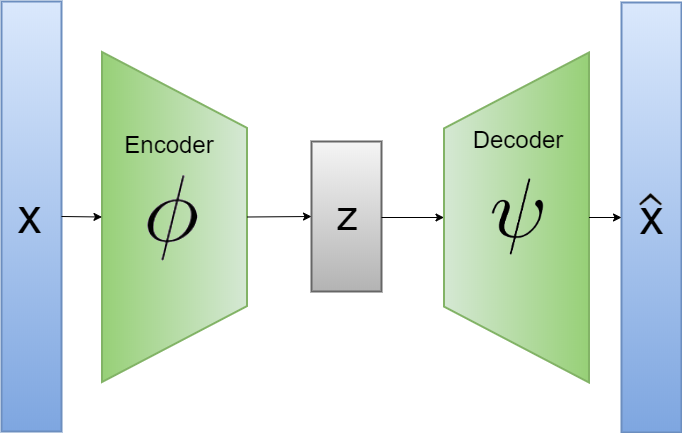}
    \caption{Autoencoder: The encoding layers $\phi$ reduce the dimension of $x$ to $z$ and the decoding layers $\psi$ project the data back to the original dimensions resulting in $\hat{x}$}
    \label{fig:Autoencoder}
    \end{figure}
    There have been different approaches using autoencoders to detect anomalies on spatial data. \citea{AnomalyDetectionRobustDeepAutoencoders} use robust autoencoders to detect anomalies on images. \citea{DeepAutoencodingModelsBrainImages} use deep autoencoders to detect anomalies on 2D brain MR images. \\
    \citea{AnomalyDetectionAutoencodersNonlinearDimensionalityReduction} use an autoencoder to detect anomalies on time-series and compare it with linear and kernel PCA. They implement a normal autoencoder and a denoising autoencoder. Denoising autoencoders contaminate the input $X$ with some noise and try to reproduce the noise-free input. Here in this paper, we will implement the normal autoencoder to detect anomalies on time-series. Thus, let $\{X_N\}$ be a univariate time-series and $w$ be the width of the sliding window on our time-series, then the input for the autoencoder will be a vector $(x_i,x_{i+1},...,x_{i+w})^T \in \{X_N\}$. Then the autoencoder will compute the following:
    \begin{equation}
        (\hat{x}_i,\hat{x}_{i+1},...,\hat{x}_{i+w})^T = \psi(\phi((x_i,x_{i+1},...,x_{i+w})^T))
    \end{equation}{}
    Using the training set, the autoencoder tries to minimize the error using the following optimization function:
    \begin{equation}
        \min_{\theta_{\psi},\theta_{\phi}}\norm{(x_i,x_{i+1},...,x_{i+w})^T- (\hat{x}_i,\hat{x}_{i+1},...,\hat{x}_{i+w})^T}_2, \forall i\in \{i,...,N-w\}
    \end{equation}{}
    Then, the test set can be used to detect the anomalies. Let $f_{\psi\phi}$ be the trained autoencoder, for an $x_j\in \{X_{TEST}\}$ we first make a prediction:
    \begin{equation}
        \hat{x}_j = f_{\psi\phi}(x_j) = \psi(\phi(x_j))
    \end{equation}{}
    After that, the error value $e_j$ for the prediction of $x_j$ will be computed:
    \begin{equation}
        e_j = \norm{x_j - \hat{x}_j}
    \end{equation}{}
    Last but not least, having an anomaly threshold $\delta \in \mathbb{R}$, $x_j$ is marked as abnormal if and only if $e_j>\delta$. As described in the previous approaches, the training data is used to determine an appropriate value for $\delta$.\\
    Autoencoders are best suited to semi-supervised learning approaches where the training data only consists of normal points. This results in learning a latent space of the normal data points and resulting in deviations when later an anomaly is fed into the model. 
\end{enumerate}{}

\section{Approach} 
\label{chapter:Approach}
In this section, we first introduce related works regarding anomaly detection. After that, the different datasets used for the evaluations are mentioned. Additionally, the data preprocessing tasks are explained. At the end, the evaluation metrics applied are listed.

%\todo[inline]{es ist immer unschön, wenn zwei überschriften direkt hintereinander kommen. hier kannst du kurz ausführen, was alles in diesem kapitel gemacht wird.}
\subsection{Related Works}
Time-series analysis have been an important topic for a long time. In the last decades, while different machine learning approaches achieved noticeable progress in various areas, there has been much effort to benefit from them in time-series analysis. There have been a lot of published works using machine learning to make better prediction. The Makridakis Competitions, also known as M-Competitions, are taking place regularly to evaluate different forecasting methods \cite{AccuracyExtrapolationMethodsResults}. Afterwards, there have been attempts to compare the best performing approaches. \citea{StatisticalMachineLeanringForecastingMethodsConsernsWaysForward} published an article where they compared the best performing statistical forecasting methods of the M-3 Competition with different machine learning approaches \cite{M3CompetitionResultsConclusionsImplications}. Their paper focuses on univariate time-series models.%\todo[inline]{was ist mit der M4 competition? - Antwort: Das Paper welches ML Methoden mit statistische Methoden vergleicht, nimmt die statistischen Methoden aus der M3 Competition. Nach der M4 Competition wurde, soweit ich weiß und es wäre auch sehr unwahrscheinlich, nicht wieder ein neues Paper publiziert. Das Paper wurde ja auch nicht von Makridakis veröffentlicht.} 

There have also been different studies comparing ML methods but for a specific sort of data. For instance, \citea{ChoosingmachineLearningAlgorithmsAnomalyDetection} compare 22 ML algorithms detecting anomalies on IoT-Datasets. There are also papers where the authors compare their anomaly detection approach with different approaches.
\citea{ForecastingBehaviorMulitvariateTimeSeriesNeuralNetwork} compare their neural network approach with the ARMA model. Furthermore, there is also research, which compares anomaly detection methods, but only evaluates on a single dataset. \citea{ComparativeStudyAnomalyDetectionSchemesNetworkIntrusionDetection} compare different clustering techniques like Mining  Outliers  Using  Distance  to  the k-th Nearest Neighbor, the Nearest Neighbor approach, Mahalanobis-distance   Based   Outlier   Detection, the LOF approach and SVM, but using just one dataset. \citea{FuseADUnsupervisedAnomalyDetectinStreamingSensorsData} compared their approach, FuseAD, with other state-of-the-art anomaly detection methods like LOF, iForest, OC-SVM, PCA, Twitter anomaly detection (TwitterAD) \cite{IntroducingPracticalRobustAnomalyDetectionTimeSeries}, and DeepAnT \cite{DeepANT_DeepLearningUnsupervisedAnomalyDetectionTimeSeries} on Yahoo Webscope dataset which is a time-series anomaly detection dataset. \\
To the best of our knowledge, there exists no extensive research comparing statistical approaches with classical machine learning approaches and neural networks detecting anomalies in univariate time-series. This is despite the fact that anomaly detection in time-series is becoming increasingly important. The most similar study to our approach is the work of \citea{M3CompetitionResultsConclusionsImplications} which compares univariate forecasting methods using statistical approaches and ML methods. \\
In this paper, we use different univariate time-series datasets to compare the univariate approaches that have been introduced in Section  \ref{chapter:chapterAnomalyDetectionApproache}. All the approaches are supervised, thus using a part of the datasets for training and other part for testing. 

% ------------------ Datasets -------------------------
\subsection{Datasets}
To evaluate the introduced univariate methods, several time-series datasets have been selected. Most of them are benchmarks for anomaly detection. One of the important criteria was that the dataset should indeed be a time-series dataset. There are different articles evaluating anomaly detection methods on time-series while choosing non-time series datasets like Forest Cover Type (ForestCover) dataset. Even the KDD Cup '99 dataset is critical, as it is not based on equal time intervals. Thus, we will only use a small portion of it to make it compatible with the time-series requirements.  \\
Therefore, several univariate datasets consisting of real and synthetic data have been chosen. Furthermore, time-series data are preferred, which are also used in other studies enabling us to compare our results with theirs when using similar methods. 
\subsubsection{UD1 \--- Real Yahoo Services Network traffic}
This dataset, which is published by Yahoo \cite{DatasetYahoo}, is a univariate time-series dataset containing the traffic to Yahoo services. The anomalies are labeled by humans. This dataset consists of 67 different time-series each containing about 1400 timestamps. The timestamps have been observed hourly. Most of the time-series are stationary and on average each time-series consists of 1420 timestamps where 1.9\% are anomalies. 

\subsubsection{UD2 \--- Synthetic Yahoo Services Network traffic}
Yahoo made another dataset \cite{DatasetYahoo} available which consists of 100 synthetic univariate time-series data containing anomalies. Each time-series contains about 1421 timestamps. The anomalies have been inserted randomly therefore representing point anomalies. On average, each time-series consists of 0.3\% anomalies. 

\subsubsection{UD3 \--- Synthetic Yahoo Services with Seasonality}
This dataset also consists of 100 synthetic univariate time-series \cite{DatasetYahoo}, each containing about 1680 timestamps. In contrast to the former one, this dataset also contains seasonality. The anomalies are inserted at random points marking the changing points. On average, the anomalous rate of each time-series is about 0.3\%. 

\subsubsection{UD4 \--- Synthetic Yahoo Services with Changepoint Anomalies}
The dataset also contains 100 synthetic univariate time-series. Each time-series has about 1680 timestamps. The difference to the former one is the fact that this dataset also contains changepoint anomalies where the mean of the time-series changes. For our evaluation, we focus on the main anomalous points and ignore distinguishing between anomalous types. On average, 0.5\% of the dataset is anomalous.

\subsubsection{NYCT \--- NYC Taxi Dataset} 
This is a univariate time-series dataset containing the New York City (NYC) taxi demand from 2014-–07--01 to 2015--01--31 with an observation of the number of passengers recorded every half hour containing 10320 timestamps. It is from the Numenta Anomaly Benchmark (NAB), which is a benchmark for evaluating algorithms for anomaly detection, especially on streaming data. It contains five collective anomalies, which occur on the NYC marathon, Thanksgiving, Christmas, New Years day, and a snow storm.

\subsection{Data Preprocessing}
\subsubsection{Standardize data}

One of the main data preprocessing tasks performed on the datasets, before evaluating the anomaly detection methods, is standardizing. A dataset is standardized, if its mean $\mu$ is zero and its standard deviation $\sigma$ is one. Thus, let $\mathcal{D}$ be the dataset and $\mu$ the mean of $\mathcal{D}$ and $\sigma$ the standard deviation. Then, to standardize $\mathcal{D}$:
\begin{equation}\label{Standardization}
 \hat{x} = \frac{x - \mu}{\sigma}, \forall x\in \mathcal{D}
\end{equation}
Standardization is not equal to normalization, where $\mathcal{D}$ is changed, such that:
\begin{equation}
    x \in [0,1], \forall x \in \mathcal{D}
\end{equation}
Normalization is sensitive to outliers. Thus, it would be inappropriate to normalize our datasets, which contain outliers. \\
Standardization in Equation \ref{Standardization} helps many ML methods to converge faster. \citea{EffectDataStandardizationNeuralnetworkTraining} have shown that especially in smaller datasets, the neural network yields better results when the data is standardized. \citea{WindSpeedForecastingBasedSVM}
mention in their paper that standardizing time-series data is necessary for the SVM approach. Also other approaches which are evaluated in this paper like DBSCAN, K-Means, and K-NN benefit from standardization. 

\subsubsection{Deseasonalizing and Detrending}
Most statistical approaches for time-series analysis presume stationarity in mean and variance. But for machine learning approaches and neural networks, this is a debatable requirement. There are studies neglecting the need for transforming the time-series into a stationary sequence. Neural networks are universal function approximators and therefore several researchers claim that they are able to detect non-stationary trends. 
Therefore, \citea{ResearchProspectiveNeuralNetwrokForecasting} believes that they are able to detect non-linear trends and seasonality in the time-series dataset. \citea{ConnectionistApproachTimeSeriesPredictionEmpiricalTest} show empirically, that neural networks are able to detect automatically the seasonality in univariate time-series by evaluating about 101 different time-series. There have also been other works achieving similar results \cite{RecognisingSeasonalPatternNN, FeedForwaredNeuralNetsTimeSeriesForecasting}.

There have also been studies achieving contradicting results.  \citea{Faraway1998TimeSF} compared models using detrended and deseasonalized time-series to networks using the origin data. They show empirically, that the model using detrended and deseasonalized data achieve better results. \citea{ZhangNNForecastingSeasonalTrendTimeseries} make an empirical study comparing the performance of neural networks using detrending and deseasonalizing performs better. In particular, they found out that the combination of both achieves the best results concluding that NN are not able to detect seasonal and trend variations effectively. \\ 
Therefore, to get more precise results when comparing different approaches, we analyze the performance of the methods using the raw time-series dataset (except for standardization) compared to detrended and deseasonalized time-series.

\subsection{Evaluation Metrics}

\subsubsection{F-Score}

To compare the different anomaly detection methods, several metrics can be taken into consideration. One of the metrics used in similar approaches is the F-Score:
\begin{equation}
    \text{F-score} = 2\cdot \frac{\text{Precision}\cdot \text{Recall}}{\text{Precision}+ \text{Recall}}
\end{equation}{}
\citea{DeepANT_DeepLearningUnsupervisedAnomalyDetectionTimeSeries} use this metric to evaluate their anomaly detection methods on different time-series. \citea{dLSTMAnomalyDetectionDeepLearning} also use the F-Measure in addition to recall and precision.

\subsubsection{Area under the curve (AUC)}

Another metric that is often used is the \textit{receiver operating characteristic curve}, ROC-Curve, and the associated metric \textit{area under the curve (AUC)}, which is the area under the ROC-Curve. This measure is very helpful, especially for anomaly detection. The ROC-Curve illustrates the correlation of the \textit{true positive rate} and the \textit{false positive rate} based on different threshold values. The true positive rate (TPR) is defined as follows:

Let P be the positive labeled values, e.g., timestamps which are actually anomalous, N the negative labeled values, e.g., timestamps which are actually normal, TP be the true positive classifications, i.e., timestamps which are anomalous and have been detected as anomalous by the algorithm, FP the false positive, i.e., timestamps which are normal but have been labeled falsely as anomalies, then TPR is:
\begin{equation}
    TPR = \frac{TP}{P}
\end{equation}
and FPR is:
\begin{equation}
    FPR = \frac{FP}{N}
\end{equation}{}
To compute the ROC-Curve, we use different $\delta \in \mathbb{R}$ as threshold for our anomaly detection method resulting in different pairs of TPR and FPR for each $\delta$. This values can finally be plotted showing a curve starting at the origin and ending in the point (1,1). The associated metric AUC is the area under the curve. In anomaly detection, the AUC expressed the probability that the measured algorithm assigns a random anomalous point in the time-series a higher anomaly score than a random normal point. This makes AUC appropriate to compare the different anomaly detection methods.  Several other papers \cite{AnomalyDetectionAutoencodersNonlinearDimensionalityReduction,LongShortTermMemoryNetworkAnomalyDetectionTimeSeries,UnsupervisedNoveltyDetectionUsingDeepAutoendcodersClustering,ComparativeStudyAnomalyDetectionSchemesNetworkIntrusionDetection} also used the ROC-Curve and AUC-Value to compare different anomaly methods. For instance,  \citea{ComparativeEvaluationUnsupervisedAnomalyDetectionMultivariateData} used this measure to compare different machine learning anomaly detection methods. 

In a nutshell, the AUC-value will be the main measure used in the experiments of this paper in Section \ref{chapter:results}.

\subsubsection{Computation Time Comparison}
Another factor to evaluate the different approaches, is the computation time a method uses to analyse the data. The time an algorithm needs to predict if a timestamp is abnormal or not plays an essential role in the success of the approach. There are many use cases where the inference time is crucial. For instance, to detect online anomalies the approach must be able to respond quickly if new data points appear rapidly.\\
To measure the computation time, different approaches are possible, based on the nature of the algorithms. For instance,  \citea{dLSTMAnomalyDetectionDeepLearning} computed the time used for training and inference separately.  \citea{ComparativeEvaluationUnsupervisedAnomalyDetectionMultivariateData} computed the time the unsupervised anomaly detection methods needed to analyse a dataset.

The method to measure the time performance in this paper is different to the approaches used in the aforementioned papers. This is based on the fact, that those papers compare similar detection methods. For instance,  \citea{ComparativeEvaluationUnsupervisedAnomalyDetectionMultivariateData} just compares machine learning approaches, especially clustering and density approaches. Also, \citea{dLSTMAnomalyDetectionDeepLearning} compares deep learning methods making it possible to measure the training and inference phase separately. This is not possible when comparing deep learning methods like LSTM with a classical machine learning method like LOF. While LSTM mostly needs a big amount of time in training, its inference time is very fast. This is completely different to clustering methods such as LOF.\\
Hence, the time performance of the algorithm is mainly compared based on the total training and prediction time. This will provide a relative value to compare the approaches with each other. 

\section{Experiments}
\label{chapter:Experiments}
In this section, the settings for the experiments are listed. This is necessary to evaluate the results of the different approaches. 

\subsection{Datasets}
The datasets are divided into training and test set where $30\%$ are used for training and $70\%$ as test data. If the test data does not contain any anomalous point, the time-series will be ignored and excluded from the evaluation. Additionally, the data is standardized before being used by the methods as explained in Section \ref{chapter:Approach}. The test data forms the source of the method evaluation. The AUC-Value is computed on the test data while the train data is only used for training. For some deep learning methods, $10\%$ of the training data is used as validation set to optimize the hyperparameters. 

\subsection{Experimental Setup}
The main requirement for the evaluation is achieving an optimally trained model for each anomaly detection method. This requires to optimize the hyperparameters of the model in addition to the trainable parameters. Therefore, a validation set is used. This prevents overfitting in the training process. \\
In addition to that and especially for the anomaly detection methods, which are based on a forecasting model, hyperparameter tuning is performed on the forecasting model. An optimized forecasting model will achieve better results later in detecting anomalies. To evaluate the forecasting model, a \textit{naive model} is created. In some literature this naive model is also called \textit{persistent model}. The naive model in time-series forecasting is a model where the output $\hat{x}_t$ of the model is equal to the input:
%\begin{equation}
%        \hat{x}_t = x_{t-1} = f_{Naive Model}((x_{t-w},...x_{t-1}))
%\end{equation}{}
\begin{equation}
    \begin{split}
         \hat{x}_t = & f_{Naive Model}((x_{t-w-1},...,x_{t-1}))\\
         where \text{ } & f_{Naive Model} : \mathbb{R}^w \to \mathbb{R}\\
          & f_{Naive Model}((x_{t-w-1},...,x_{t-1})) \mapsto x_{t-1}
        % \phi : \mathbb{R}^N & \to \mathbb{R}\\
        % \phi(x) & \mapsto \gamma 
 \end{split}{}
\end{equation}{}
The accuracy of the naive model forms a lower bound for the target model. Following this, the Mean Squared Error (MSE) of the target model should be lower than the naive model. Therefore, this paper suggests the following metric to evaluate the forecasting performance of the prediction model:
\begin{equation}
    \text{NMM}(MSE_{Target model},MSE_{Naive Model}) = \frac{MSE_{Target model}}{MSE_{Naive Model}}
\end{equation}
where NMM stands for \textit{Naive Model Metric}. Empirically, we have noticed that a model usually has a higher AUC-Value, when the NMM is lower. This has been especially useful when the dataset is unlabeled or the rate of anomalies in the training set is extremely low. Therefore, the hyperparameters of the model are tuned to minimize the NMM value aiming to keep it strictly lower than 1. 

\subsubsection{Involved Software and Hardware}
\paragraph{Software}
The algorithms are implemented in python. For the statistical approaches, mainly the module \textbf{Statsmodels} \cite{seabold2010statsmodels} was used, while some custom algorithms like PCI were implemented from scratch. The classical machine learning approaches have been mostly implemented using \textbf{Scikit-learn (Sklearn)} \cite{scikit-learn} and for the deep learning approaches the \textbf{tensorflow} \cite{tensorflow2015-whitepaper} and \textbf{keras} library \cite{chollet2015keras} have been used. \\
\paragraph{Hardware}
The training and testing of the models were performed on the hardware listed in Table \ref{tbl:HardWareSpecification}:

\begin{table}[H]

\centering
\begin{tabular}{ll} 
\toprule
Artifact        & Value                                                 \\ 
\midrule
CPU Model name: & 2xIntel(R) Xeon(R) CPU @ 2.30GHz, 46MB Cache, 1 Core  \\
RAM             & \textasciitilde{} 12.4 GB                             \\
GPU             & 1xTesla K80, 12GB                                     \\
\bottomrule
\\
\end{tabular}
  \caption{Hardware specification}
  \label{tbl:HardWareSpecification}
\end{table}
All computations have been performed on a single process and a single thread.

\newpage

\subsubsection{Hyperparameter Tuning}\label{HyperparameterTuning}
In the following, we describe the general and the specific hyperparameters for the evaluation and the respective methods.

\paragraph{General Hyperparameters}
Some of the hyperparameters are general and dependent on the dataset, unless another value is listed explicitly for a specific approach. Table \ref{TblGeneralHyperparameters} lists these hyperparameters:

\begin{table}[H]
\begin{tabular}{   L{8cm}  L{4cm}  L{3cm} }
\toprule
Dataset                        & Hyperparameter       & Value  \\ 
\midrule
Yahoo Services Network traffic  & Sliding window width $w$ & 30     \\
\midrule
All datasets (NYCT and UD1 - UD4)                  & Ratio of training set                  &  0.3      \\
                                & Ratio of test set          & 0.7\\
                                & Ratio of validation set & 0.1 of test set\\
\bottomrule
\\
\end{tabular}
  \caption{General Hyperparameters} \label{TblGeneralHyperparameters}
\end{table}

\paragraph{Statistical approaches}
Table \ref{tbl:HyperparameterStatistical} lists the hyperparameters of the statistical approaches:
    \begin{table}[H]
    \centering
    \resizebox{\textwidth}{!}{%
    \begin{tabular}{   L{4cm}  L{4cm}  L{7cm} }
    \toprule
    \multicolumn{3}{c}{\textbf{Statistical Approaches}} \\
    \toprule
    Model & Hyperparameter       & Value                                           \\ 
    \midrule
    AR-Model & maximal lag    & $12\cdot(\frac{|X_{Train}|}{100})^{\frac{1}{4}}$  \\
     & Fitting method & Conditional maximum likelihood using OLS        \\        \midrule
    MA-Model &maximal lag    & $12\cdot(\frac{|X_{Train}|}{100})^{\frac{1}{4}}$  \\
     & error residual lag    & Sliding window width $w$  \\
    & Fitting method & Conditional maximum likelihood using OLS        \\
    \midrule
    ARIMA-Model &maximal lag    & $12\cdot(\frac{|X_{Train}|}{100})^{\frac{1}{4}}$  \\
     & $p$    & 1   \\
     & $d$    & 1 if data contains trend, otherwise 0\\
     & $q$    & 2 \\
    & Fitting method & Maximizing Conditional Sum of Squares likelihood       \\
    & Maximal iteration & 500\\
    & Convergence tolerance & $10^{-8}$\\
    \midrule
    SES  & Smoothing Parameter $\alpha$ & Use grid search to find the best values using MLE\\
    \midrule
    ES  & $\alpha,\beta,\gamma$ & Use grid search to find the best values using MLE \\
    \midrule
    PCI             & $k$             & 30 \\
                    & $\alpha$      & 98.5 \\
    \bottomrule
    \\
    \end{tabular}}
      \caption{Hyperparameters of the statistical approaches}
      \label{tbl:HyperparameterStatistical}

    \end{table}

\paragraph{Machine Learning approaches}
Table \ref{tbl:HyperparameterML} lists the hyperparameters of the machine learning approaches:   

    \begin{table}[H]
    \centering
    \begin{tabular}{   L{4cm}  L{4cm}  L{7cm} }
    \toprule
    \multicolumn{3}{c}{\textbf{Machine Learning Approaches}} \\
    \toprule
    Model & Hyperparameter       & Value                                           \\ 
    \midrule
    K-Means & k    & 4   \\
     \midrule
    DBSCAN  & $\epsilon$    & 0.4   \\
            & $\mu$         & 5     \\
            & distance function & Euclidean distance\\
    \midrule
    LOF     & $k    $       & 10    \\
            & distance function & Minkowski distance\\

    \midrule
    Isolation Forest & Number of iTrees & 10 \\
    (iForest)            & Contamination value & Find automatically\\
    \midrule
    One-Class SVM (OC-SVM)   & kernel       & Radial basis function kernel(RBF)\\
            & Upper bound of outliers & $0.7$\\
    \midrule
    XGBoosting & Max Tree depth & 3             \\
    (XGB)            & Learning rate & 0.1           \\
                & Number of estimators & 1000   \\
                & Loss function & MSE           \\
    \bottomrule
        \\
    \end{tabular}
      \caption{Hyperparameters of the Machine Learning approaches}
      \label{tbl:HyperparameterML}
    \end{table}
 
\paragraph{Deep Learning approaches}
Table \ref{tbl:HyperparameterDL} lists the hyperparameters of the deep learning approaches:
    \begin{table}[H]
    \centering
    %\resizebox{\textwidth}{!}{%
    \begin{tabular}{   L{3cm}  L{4cm}  L{8cm} }
    \toprule
    \multicolumn{3}{c}{\textbf{Deep Learning Approaches}} \\
    \toprule
    Model & Hyperparameter       & Value                                           \\ 
    \midrule
    MLP & Number of hidden layers    & 2                            \\
        & Neurons in each hidden layer & 100, 50                    \\
        & Activation function       & ReLU                          \\
        & Optimizer, Loss           & Adam, MSE                     \\
        & Batch size,Epochs          & 32, 50                        \\
    \midrule
    CNN    & Architecture              & 3 Convolution Blocks with Max-Pooling and ReLU, \\
        &                           & then one Dense layer with 50 neurons and ReLU\\
        & Filters                   & 8,16,32 with kernel size 2     \\
        & Optimizer, Loss                 & Adam, MSE               \\
        & Batch size, Epochs         & 32, 50                        \\
    \midrule
    CNN+Batch       & Architecture &  2 Convolution Blocks with Batch-Normalization with ReLU\\
    Normalization   & & then one Dense layer with 50 neurons and ReLU\\
    (CNN\_B)         & Filters       & 256,256 with kernel size 3\\
                    & Optimizer, Loss & Adam, MSE\\
                    &Batch size, Epochs & 32,50\\
    \midrule
    CNN+Residual& Architecture & 1 Convolution Block with One Residual Block \\
    Blocks      &               & then one Dense Layer with 50 neurons and ReLU\\
    (CNN Residual) & Filters       & 256,256,256 with kernel size 3\\
                & Optimizer, Loss & Adam, MSE\\
                & Batch size, Epochs & 32, 50\\    

    \midrule
    \end{tabular}
    \end{table}

    \begin{table}[H]
    \centering
    \resizebox{\textwidth}{!}{%
    %\resizebox{\textwidth}{!}{%
    \begin{tabular}{   L{3cm}  L{4cm}  L{8cm} }

    \midrule
    WaveNet     & Architecture & 3 Convolution Blocks with Max-Pool and ReLU function,\\
                &               & with Dilation rate 1,2,4 then one Dense layer with\\  
                &               &  50 neurons and ReLU\\    
                & Filters       & 8, 16, 32\\
                & Optimizer     & Adam, MSE\\
                & Batch size, Epochs & 32, 50\\    
    \midrule
    LSTM        & Architecture  & 2 stateful LSTM Layer \\
                & Filters       & 4,4\\
                & Optimizers, Loss & Adam, MSE\\
                & Batch size, Epochs & 32, 50\\                
    \midrule
    GRU         & Architecture  & 2 stateful LSTM Layer \\
                & Filters       & 4,4\\
                & Optimizers, Loss & Adam, MSE\\
                & Batch size, Epochs & 32, 50\\
    \midrule

    Autoencoder & Architecture  & 2 Encoding Layers (32,16), 2 Decoding Layers (16,32)  \\             & Activation functions & ReLU for Decoding and Encoding, linear for output\\
                & Optimizer, Loss & Adam, MSE\\
                & Batch size, Epochs & 32, 50\\
    \bottomrule
    \\
    \end{tabular}}
      \caption{Hyperparameters of the Deep Learning approaches}
      \label{tbl:HyperparameterDL}
    \end{table}

%---------------------------- RESULTS ----------------------------
\section{Results}
\label{chapter:results}
In this Section, the results of the different approaches are presented. The approaches from Section \ref{chapter:chapterAnomalyDetectionApproache} with the hyperparameters listed in Section \ref{chapter:Experiments} have been executed. The following sections list the AUC-Values and computation time, respectively. 

\subsection{AUC-Values}
Table \ref{tbl:AUC-ValuesAll} lists the AUC-Values obtained on the different datasets by the univariate approaches. The AUC-Value was computed on each dataset separately: 
\begin{table}[H]
\centering
\resizebox{\textwidth}{!}{%
    \begin{tabular}{   L{1.1cm}   L{1.5cm}  L{1.5cm} L{1.5cm}  L{2.2cm}  L{1.5cm} L{1.5cm} L{1.2cm}  L{1.9cm} }
\toprule
Dataset & AR & MA & SES & ES & ARIMA  & PCI 	  \\
\midrule
UD1 & \textbf{0.911394} & 0.868123 & 0.824894 & 0.830289 &  0.8730  & 0.522397  		 \\
UD2 & 0.994769 & 0.994150 & 0.932215 & 0.957964 &  0.9891  & 0.762529  		 \\
UD3 & 0.994116 & \textbf{0.994245} & 0.990782 & 0.989360 &  0.990  & 0.674337  		 \\
UD4 & 0.975152 & \textbf{0.986400} & 0.969333 & 0.971991 &  0.9709  & 0.688257  		 \\
NYCT &  0.6369 & 0.3938 & 0.3452 & 0.3423 & 0.3583 & 0.5377  \\
\midrule
Dataset &K-Means & DBSCAN 	& LOF 		& iForest 	& OC-SVM & XGBoost	& 		\\
\midrule		
UD1 & 0.877623 & 0.806574	& 0.814574 & 0.803997 	&0.850292 & 0.896743	& 	\\
UD2 & 0.923446 & 0.995251	 & 0.995116 & 0.993984  &\textbf{0.995276} & 0.968308	& \\
UD3 & 0.728037 & 0.696868	 & 0.951007 & 0.952241  &0.957272 & 0.80231 	& \\
UD4 & 0.663028 & 0.725566	 & 0.952523 & 0.955123  &0.939444 & 0.85375 	& \\
NYCT & \textbf{0.9137} & 0.5407 & 0.5294 & 0.4922 & 0.5859 & 0.4602 &  \\

\midrule
Dataset & MLP & CNN & CNN B & CNN Residual & Wavenet& LSTM & GRU & Autoencoder  \\
\midrule
UD1 &    0.780471 & 0.809449 & 0.785505 & 0.819109 & 0.8239	 & 0.8121 & 0.8025 & 0.782197       \\
UD2 &    0.71911 & 0.74924 & 0.74815 & 0.721279 & 0.761423	 & 0.7348 & 0.7183& 0.742767      \\
UD3 &   0.569883 & 0.582761 & 0.581911 & 0.575338 & 0.579589 & 0.5781 & 0.5711& 0.602905	    \\
UD4 &   0.558124 & 0.581019 & 0.604943 & 0.568564 & 0.592414 & 0.5891 & 0.5811 & 0.597238	    \\
NYCT & 0.7962   & 0.8181    & 0.76      & 0.7468    & 0.8229 & 0.8404 & 0.7978 & 0.6967 \\ 
    \bottomrule
    \\
    \end{tabular}}
  \caption{AUC-Values on each dataset by each approach} 
  \label{tbl:AUC-ValuesAll}
\end{table}

The best results are highlighted in table \ref{tbl:AUC-ValuesAll}. To provide a better illustration of the results, the average AUC-Value of all datasets UD1-UD4 for each algorithm is computed and plotted as a sorted sequence in figure \ref{fig:AucValue}:

\begin{figure}[H]
 \centering
  \captionsetup{justification=centering,margin=2cm}
 \includegraphics[scale=0.6]{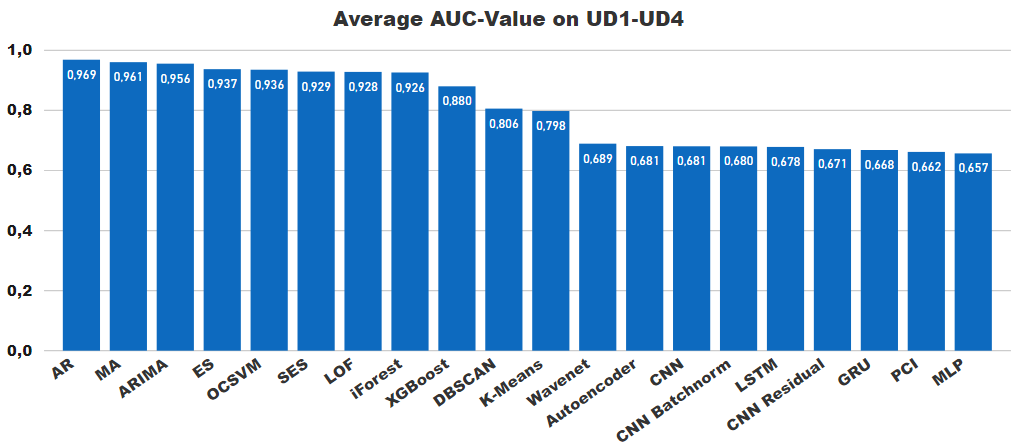}
 \caption{Average AUC-Value computed on UD1-UD4}
 \label{fig:AucValue}
\end{figure}

Figure \ref{fig:AucValue} shows that the statistical models achieved the best results while the deep learning methods generally performed poorly. Four of the five best performing algorithms are statistical while four of the worst performing algorithms are deep learning approaches. Most of the machine learning approaches are located in the center. Only PCI is an exception of the statistical approaches, which reached a very low AUC-value.\\
However, surprisingly the deep learning methods perform much better on the NYCT dataset while the statistical approaches achieve a very low AUC-Value. Figure \ref{fig:NYCTAucValue} shows how the different approaches behave on the NYCT dataset:
%\todo[inline]{this hardly qualifies as a sentence}:
\begin{figure}[H]
 \centering
  \captionsetup{justification=centering,margin=2cm}
 \includegraphics[scale=0.6]{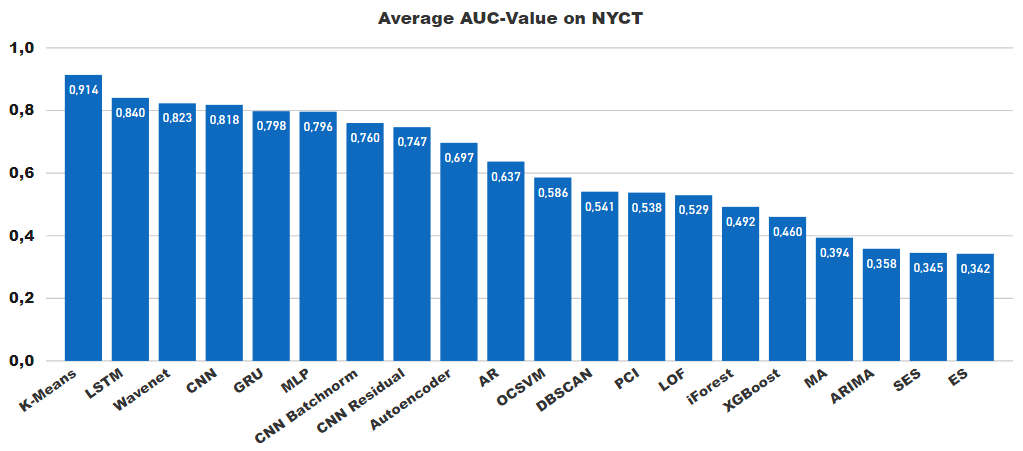}
 \caption{Average AUC-Value computed on UD1-UD4}
 \label{fig:NYCTAucValue}
\end{figure}
The main reason to this poor performance of the statistical approach is the fact that the NYC time-series contains contextual anomalies while the anomalies in UD1, UD2, UD3 and UD4 are either point anomalies or collective anomalies. Most of them do not contain any contextual anomalies. The values of the anomalous points differ clearly from the rest normal points. But in the NYCT dataset the value of the anomalous points is similar to the normal points. They are anomalous due to their contextual information. We observed that the statistical models overfit to these data, preventing them to detect the anomalies. On the other side, deep learning approaches provide a more flexible way to optimize the model according to the anomaly type by tuning the broad range of hyperparameters making them able to achieve high AUC values.

\subsection{Computation Time}
The computation time is measured for each algorithm on the datasets UD1-UD4 containing 367 time-series which is listed in the first line of the following table. Then the average time needed to train and detect anomalies on a single time-series is computed and listed in the second line. The time is measured in seconds.
\begin{table}[H]
\centering
\resizebox{\textwidth}{!}{%
\begin{tabular}{lllllllll} 
\toprule
Dataset & AR & MA & SES & ES & ARIMA  & PCI 	  \\
\midrule
UD1-4  			& 51 & 17 & 2910 & 13268 & 72422 & 374  		 	\\
One time-series 	& \textbf{0.139} & 0.046 & 7.93 & 36.15 & 197.3351 & 1.02 	\\

\midrule
Dataset &K-Means & DBSCAN 	& LOF 		& iForest 	& OC-SVM & XGBoost	& 		\\
\midrule		
UD1-4  			& 190  & 319 	& 319 & 38451   & 210  	& 14	&		\\
One time-series 	& 0.52 & 0.87 	& 0.87 & 104.77  & 0.57 	& 0.38	&  	\\

\midrule
Dataset         & MLP & CNN & CNN B & CNN Residual & Wavenet & LSTM & GRU & Autoencoder  \\
\midrule
UD1-4  			& 7396  & 10824 & 12938 & 37038 	& 17252 & 54083 & 49756	&  9413 \\
One time-series 	& 20.15 & 29.49 & 35.38 & 100.92 & 47.0  & 147.36 & 135.57	&  25.65\\
\midrule
\\
\end{tabular}}
  \caption{Computation Time of the different approaches on UD1-UD4 datasets} \label{tbl:ComputationTime}

\end{table}
The results in Table \ref{tbl:ComputationTime} show that the autoregression models AR and MA are the fastest algorithms while in general the deep learning methods have a huge computation time. But not all statistical approaches benefit from a lower runtime. SES and ES require more time than many machine learning approaches. Deep learning approaches have higher computation time because of their time-consuming training phase. Especially, LSTM and GRU are the bottom of the table. \\
Some papers argue that deep learning approaches invest most of their computation time in the training phase and that the inference time in deep learning approaches is much faster than other machine learning methods. But this statement is not always true as table \ref{tbl:ComputationTimeNYCT}, which contains the measured training and inference time on the NYCT dataset, demonstrates:
\begin{table}[H]
\centering
\resizebox{\textwidth}{!}{%
\begin{tabular}{lllllllll} 
\toprule
             & AR & MA & SES & ES & ARIMA  & PCI 	  \\
\midrule
Training Time	& 0.1004 & 0.1404 & 125.26 & 989.49 &  1482.87 & 2.85  		 	\\
Inference Time 	& 0.1004 & 0.6423 & 5.06 & 9.5115 & 1481.56 & 0.0093 	\\

\midrule
Dataset &K-Means & DBSCAN 	& LOF 		& iForest 	& OC-SVM & XGBoost	& 		\\
\midrule		
Training Time   & 0.25845 & 0.0879 	& 0.0917 & 5.5646   & 0.02  	& 0.1187	&		\\
Inference Time  & 1.99455 & 8.3774	& 5.5646 & 787.29  & 2.98	& 0.0205	&  	\\

\midrule
Dataset         & MLP & CNN & CNN B & CNN Residual & Wavenet & LSTM & GRU & Autoencoder  \\
\midrule
Training Time 		& 4.439  & 48.322 & 39.2339 & 111.36 &23.263 & 1067 & 813.021	&  17.459 \\
Inference Time  	& 0.561 & 0.6783 & 0.7661 & 1.6401 &0.737   & 1086  & 385.979 &  0.5406\\
\bottomrule
\\
\end{tabular}}
  \caption{Training and Inference compuatation time on the NYCT dataset} \label{tbl:ComputationTimeNYCT}
\end{table}
The training time represents the computation time we measured by fitting the model on the trainings set which consists of 30\% of the NYCT dataset. The inference time was the measured time, the model needed to label the rest of the NYCT dataset which represents the test set.\\
Most of the deep learning approaches have a very low inference time. But as the measured values in the table show, recurrent neural network like LSTM and GRU also have a very huge inference time exceeding the inference times of almost all machine learning approaches. Hence, although the LSTM model achieved the best results on the NYCT dataset, but its inference time remains critical.

\subsection{Computation Time vs AUC-Value}
\begin{figure}[H]
 \centering
  \captionsetup{justification=centering,margin=2cm}
 \includegraphics[scale=0.45]{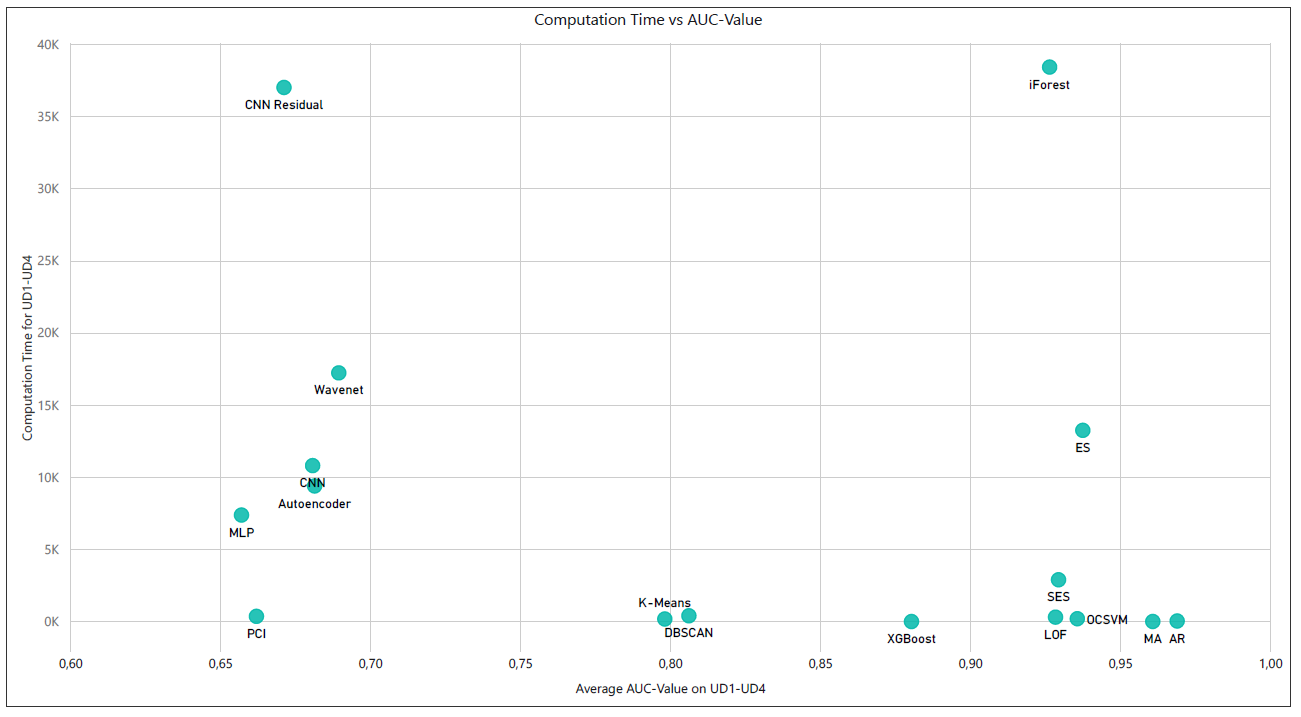}
 \caption{Average AUC-Value vs computation time}
 \label{fig:AucValueVsComputationTime}
\end{figure}

An interesting relation is the accuracy a method can achieve compared to the computation time it requires. Figure \ref{fig:AucValueVsComputationTime} displays how the different approaches perform in regard to AUC-Value and Computation time on the datasets UD1-UD4. The best performing algorithms would locate at the lower right part of the graph having a high AUC-Value and a low computation time. The graph shows that the statistical methods are performing best, i.e., AR and MA model. On the other hand, most deep learning methods are showing low AUC-Value. CNN with residual blocks is performing worse than any other method having low time performance and low AUC-Value at the same time.

\subsection{Findings}
The evaluation on the univariate time-series data has shown that despite the advances in machine learning approaches and deep neural networks, still the statistical methods that rely on the generating process are generally performing best. In addition to the fact that these methods detect anomalies more accurately than the other approaches, they also perform faster and have a very low training and prediction duration. Furthermore, the optimization of the statistical methods is much easier compared to deep learning methods, because the number of hyperparameters of the statistical algorithms is low. The results have also highlighted the fact that in time-series anomaly detection the machine learning approaches usually perform better than the deep learning methods, although the computation time of several of them is critical. For instance, the isolation forest has a very high AUC-value, but needs a thousand times more time for training and inference than the AR and MA model. This is despite the fact that the number of trees selected in our experiments have been quite moderate. But at the same time, the experiments have also shown that statistical approaches have trouble dealing with contextual anomalies. In cases like this, neural network approaches and some machine learning approaches could achieve notably higher AUC-values. This finding is important, because a similar comparison of statistical approaches and machine learning approaches for forecasting univariate time-series concluded that statistic approaches dominated the machine learning and deep learning approaches across in both accuracy and for all forecasting horizons they examined \cite{StatisticalMachineLeanringForecastingMethodsConsernsWaysForward}. Hence, these findings show how anomaly detection on univariate time-series data differ from exclusive univariate forecasting.\\

\section{Conclusion and Future Work}
    \label{chapter:Conclusion}
    In this paper a comparison between statistical and classical machine learning and deep learning approaches has been performed using a wide range of state of the art algorithms. Overall, we evaluated 20 univariate anomaly detection methods by using five univariate time-series datasets consisting of 368 univariate time-series datasets to provide a reliable contribution for the research community about the performance of these three classes of anomaly detection methods. \\ 
    The experiments showed that the statistical approaches perform best on univariate time-series by detecting point and collective anomalies. They also require less computation time compared to the other two classes. Although deep learning approaches have gained huge attention by the artificial intelligence community in the last years, our results have revealed that they are not generally able to achieve the accuracy values of the statistical methods on the univariate time-series benchmarks which only consist of point and collective anomalies. Only if the univariate dataset mainly consists of contextual anomalies, neural network could outperform the statistical methods. Although still suffering from high computation time compared to statistical approaches, some of the deep learning methods could outperform them by achieving higher AUC-rates. \\
    The main focus of this paper has been univariate time-series. However, many real data sets consist of multivariate data. Therefore, anomaly detection on multivariate time-series will be the subject of future research. Additionally, one related field, which is also of current interest, is Online Anomaly Detection. This paper focused on anomaly detection on static data, but can be extended to evaluate streaming data. \\
    In this paper, we evaluated the anomaly detection methods on general time-series datasets and the results have shown that the property of the data affects the performance of the algorithms. Therefore, it would be a line of future works to extend the experiments of this paper by evaluating the approaches on different domain-specific datasets.\\
    Finally, several attempts can be made to extend the results of this paper. However, to the best of our knowledge, this is the first attempt to provide a broad evaluation of different anomaly detection techniques on time-series data. We hope that this paper and the corresponding experiments aid other researchers in selecting an appropriate anomaly detection method when analysing time-series.   \\

%\bibliographystyle{ieeetr}
%\bibliography{references}
\bibliographystyle{plainnat}
\bibliography{AnomalyDetection.bib}
% \printbibliography
\end{document}